\documentclass{article}
\pdfoutput=1

\usepackage[preprint]{neurips_data_2023}

\usepackage[utf8]{inputenc} 
\usepackage[T1]{fontenc}    
\usepackage{hyperref}       
\usepackage{url}            
\usepackage{booktabs}       
\usepackage{amsfonts}       
\usepackage{nicefrac}       
\usepackage{microtype}      
\usepackage{xcolor}         
\usepackage{xspace}
\usepackage{graphicx}
\usepackage{amsthm}
\usepackage{latexsym}
\usepackage{amssymb}
\usepackage{multirow}
\usepackage{makecell}
\usepackage{array}
\usepackage{mathrsfs}
\usepackage{subfigure}
\usepackage{bbding}
\usepackage{pifont}
\usepackage{float}
\usepackage{CJKutf8}
\usepackage{tabularx}
\usepackage{wrapfig}
\usepackage{color}
\usepackage{soul}
\usepackage{natbib}
\setcitestyle{numbers,square}

\usepackage[capitalize]{cleveref} 
\crefname{section}{§}{§§}
\newcommand{\TVD}{\texttt{PTVD}\xspace}

\definecolor{myblue}{HTML}{d5e5ff}
\definecolor{myred}{HTML}{f08072}

\usepackage{xcolor}
\usepackage{pbox}
\usepackage[normalem]{ulem} 
\newcommand\hlb{\bgroup\markoverwith
  {\textcolor{myblue}{\rule[-.5ex]{2pt}{2.5ex}}}\ULon}
 \newcommand\hlr{\bgroup\markoverwith
  {\textcolor{myred}{\rule[-.5ex]{2pt}{2.5ex}}}\ULon}

\title{\TVD: A Large-Scale Plot-Oriented Multimodal\\ Dataset Based on Television Dramas}

\author{%
  Chen Li$^{1}$\thanks{Equal contribution}\hspace{0.4cm}Xutan Peng$^{3*}$\hspace{0.4cm}Teng Wang$^{1*}$\hspace{0.4cm}Yixiao Ge$^1$\hspace{0.4cm}Mengyang Liu$^2$\\\textbf{Xuyuan Xu}$^2$\hspace{0.4cm}
  \textbf{Yexin Wang$^2$\hspace{0.4cm}Ying Shan$^1$}\\
  $^1$ARC Lab, Tencent PCG \hspace{0.4cm} $^2$AI Technique Center of Tencent Video, Tencent PCG\\
  $^3$The University of Sheffield\\
  \texttt{\{palchenli,wybertwang,yixiaoge,myleonliu,evanxyyu},\\
  \texttt{yexinwang,yingsshan\}@tencent.com} \hspace{0.4cm}
\texttt{\{x.peng\}@shef.ac.uk}
}

\begin{document}

\maketitle

\vspace{0.1in}

\begin{abstract}
Art forms such as movies and television (TV) dramas are reflections of the real world, which have attracted much attention from the multimodal learning community recently. However, existing corpora in this domain share three limitations: (1)~annotated in a scene-oriented fashion, they ignore the coherence within plots; (2)~their text lacks empathy and seldom mentions situational context; (3)~their video clips fail to cover long-form relationship due to short duration. To address these fundamental issues, using 1,106 TV drama episodes and 24,875 informative plot-focused sentences written by professionals, with the help of 449 human annotators, we constructed \TVD, the first \textit{plot-oriented} multimodal dataset in the TV domain. It is also the first non-English dataset of its kind. Additionally, \TVD contains more than 26 million bullet screen comments (BSCs), powering large-scale pre-training. Next, aiming to open-source a strong baseline for follow-up works, we developed the multimodal algorithm that attacks different cinema/TV modelling problems with a unified architecture. Extensive experiments on three cognitive-inspired tasks yielded a number of novel observations (some of them being quite counter-intuition), further validating the value of \TVD in promoting multimodal research. The dataset and codes are released at \url{https://ptvd.github.io/}.
\end{abstract}

\vspace{0.1in}

\section{Introduction}
\label{sec:introduction}

Multimedia channels, e.g., cinema and television (TV), present stories through video, text, image, and audio. By watching how different characters interact with each other in various scenarios, audiences are able to broaden their horizons and experience colourful emotions, such as love, sorrow, and excitement.

As these plot-rich stories (e.g., movies and TV dramas) reflect our real world, they have long been used as devices for the development and assessment of human intelligence (especially for children) in the Cognitive Science community~\citep{dev-theory-mind,child-theory-mind,gunter2005children,bullet-screen-pyschology}. Analogously, a large spectrum of works employed movies and TV dramas to train and test multimodal algorithms~\citep{lsmdc2015cvpr,yms2018cvpr,movienet2020eccv,mad2022cvpr}. However, the results obtained from current methods are generally unsatisfactory, particularly when handling clips. They often only scratch the surface and fail to capture the plot of the cinema and TV.

\newpage

\begin{table*}[t]
\centering
\scriptsize
\begin{tabular}{lrrrrrrrr}
\hline
\multirow{2}{*}{\textbf{Dataset}} & \textbf{Total length} & \multirow{2}{*}{\textbf{Clip \#}} & \textbf{Avg clip length} & \multirow{2}{*}{\textbf{Sentence \#}} & \textbf{Avg sentence length} & \multirow{2}{*}{\textbf{BSC \#}} & \multirow{2}{*}{\textbf{Language}} \\
 & \textbf{(minute)} &  & \textbf{(second)} &  & \textbf{(word)} &  &  \\
\hline
LSMDC 16 & 8,820 & 128,085 & 4.1 & 128,118 & 9.0 & \ding{55} & \texttt{EN} \\
CMD & 76,185 & 33,976 & 2.4 & 35,681 & 14.5 & \ding{55} & \texttt{EN}\\
MoiveNet & 119,992 & \texttt{UNK} & \texttt{UNK} & \texttt{UNK} & \texttt{UNK} & \ding{55} & \texttt{EN} \\
Pororo & 1,231 & 16,066 & \texttt{UNK} & 43,394 & \texttt{UNK} & \ding{55}  & \texttt{EN} \\
SYMON & 52,103 & \texttt{UNK} & 2.2 & 683,611 & 13.1 & \ding{55} & \texttt{EN}\\
\TVD (Ours) & 39,908 & 24,875 & 63.2 & 24,875 & 36.9 & 26,438,007  & \texttt{ZH}\\
\hline
\end{tabular}
\caption{\label{tab:compare}
Comparisons between \TVD and related datasets. In \TVD, ``sentence'' refers to pieces of plot text. 
}
\vspace{-0.3in}
\end{table*}

\begin{wrapfigure}{r}{0.6\textwidth}
\centering
    \includegraphics[width=0.55\textwidth]{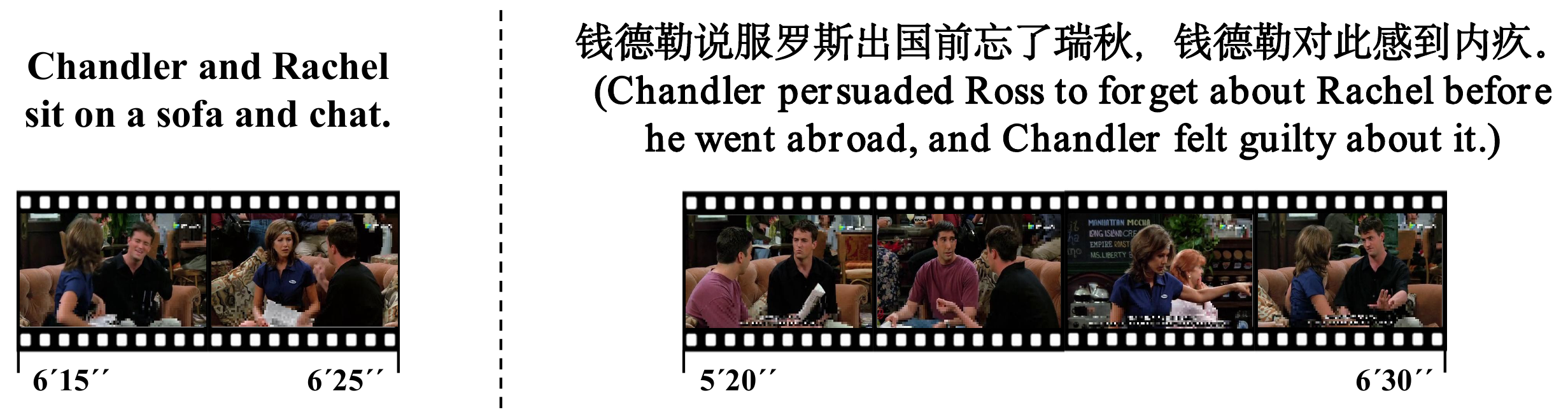}
    \caption{
    Illustrated comparisons between a traditional scene-oriented dataset (left) and \TVD (right).
    }
    \label{fig:example}
\vspace{-0.1in}
\end{wrapfigure}

Despite these prior attempts having harvested many valuable insights, the datasets they used exhibit three major problems. \textbf{First and foremost}, to the best of our knowledge, in \textit{all} these corpora segmentation is made based on scenes or shots (i.e., continuous views filmed by one camera without interruption), as illustrated in Fig.~\ref{fig:example}. While such a setup does help reduce the annotation burden, it inevitably breaks the coherence of a plot (an interrelated sequence of events in a story). It has been verified that children at the age of 5 are able to understand isolated scenes of a TV drama, but they have difficulty putting pieces of a plot together~\citep{y5-scene-plot}. Therefore, traditional datasets with scene-oriented annotations may not be sufficient to study a multimodal framework's capability of processing complex stories.
\textbf{Second}, the text signals (e.g., captions) in existing datasets are more simple descriptions of visual targets than vivid narratives of the storyline, as information such as the mental states of roles and situational contexts of events is seldom mentioned. From the perspective of Psycholinguistics, the lack of these elements indicates a deficiency of empathy and sensory perception~\citep{dev-theory-mind,child-theory-mind}. Therefore, conventional datasets are arguably limited in terms of reflecting the machine's utility in the \textit{human world}. 
\textbf{Third}, shown in Tab.~\ref{tab:compare}, clips in most existing corpora are rather short (no more than five seconds on average). As a result, they are unlikely to cover long-form relationships, thus being unsuitable for the development of more advanced systems~\citep{pretrain32022arxiv}.

To bridge the above research gaps, we construct a new dataset (namely \TVD) using large-scale TV drama resources and accompanying Chinese text.
Concretely speaking, \TVD contains 1,106 complete episodes from 83 TV drama series, with comprehensive metadata available. The total video length is 39,908 minutes.
Based on 24,875 professionally written sentences that provide empathic and context-aware plot descriptions (crawled from the web, see~\cref{sec:construction}), 449 human annotators produced 24,875 clips (each corresponding to a standalone plot that links multiple scenes). With an average duration of over 60 seconds, these plot-oriented clips mitigate the aforesaid issues of scene-oriented counterparts and are suitable to assess whether a tested system can model long-form relationships in stories.
Moreover, in \TVD we introduced Bullet Screen Comments (BSCs), a type of real-time comment generated by the audience while watching videos, to provide additional text signals. \TVD's final BSC battery consists of an astonishing number of 26,438,007 sentences, which is also the largest BSC corpus to date. All involved modalities are made parallel with aligned timestamps.

\begin{figure*}[t]
\centering
\includegraphics[width=1.0\textwidth]{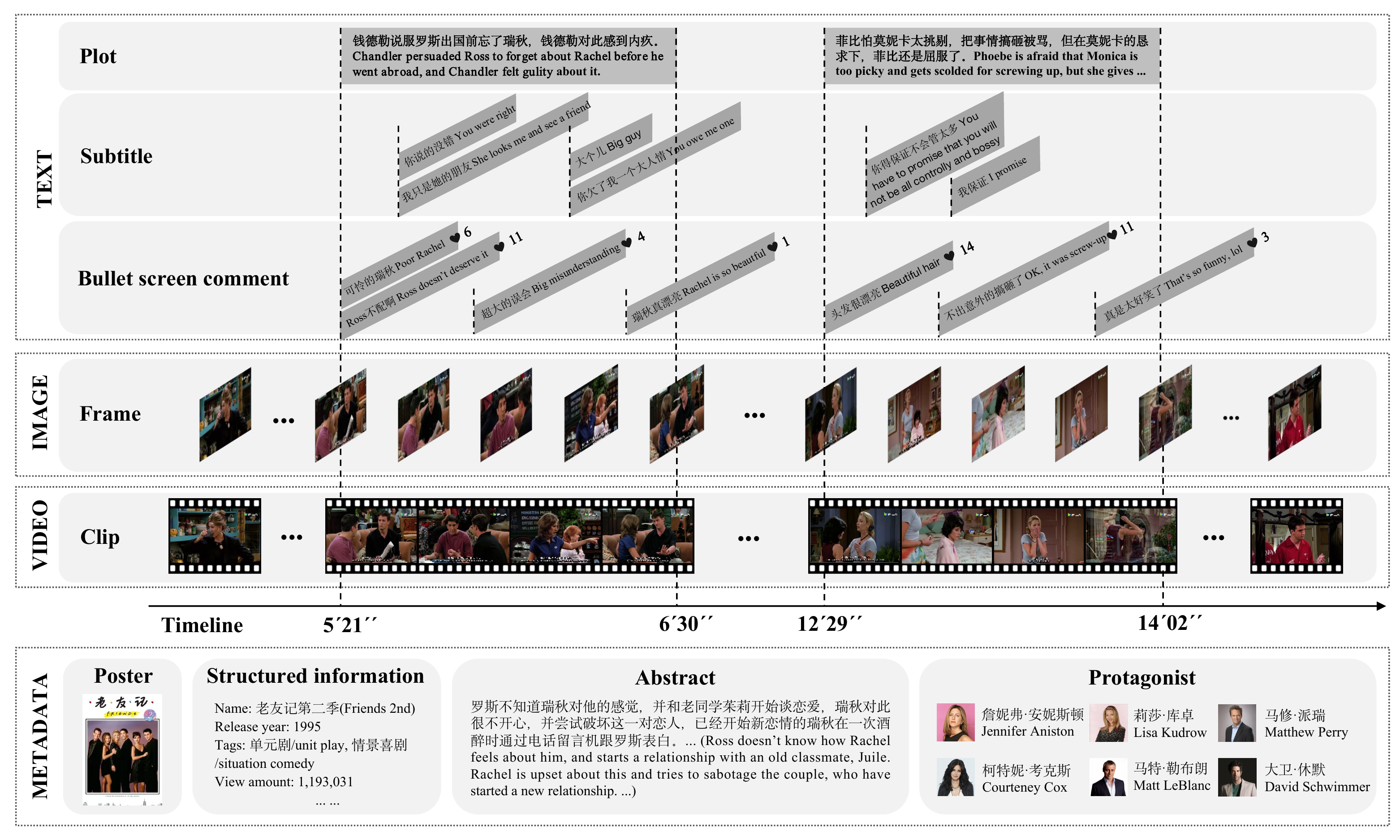}
\caption{
Data structure of \TVD. Sampled from ``\textit{Friends 2nd}'' (Ep. 1), this illustration covers two standalone plots.
}
\label{fig:dataview}
\vspace{-0.2in}
\end{figure*}

Empirically, to demonstrate how \TVD can be employed to boost multimodal research, motivated by the practice and findings in Cognitive Science, we made large-scale experiments on three tasks: Genre Classification, Plot Retrieval, and Plot Text Generation. Note that each task relates closely to one milestone of human intelligence growth (from naive to mature) (see~\cref{sec:tasks}). By adapting state-of-the-art architectures~\citep{meter2022cvpr}, we implemented a unified framework that can be further fine-tuned to solve all three tasks, which naturally serves as a strong baseline for follow-up studies.
On the one hand, we identified the weaknesses of current multimodal algorithms in the cognitive-inspired tasks of different levels.
On the other hand, through extensive ablation studies, we made a list of previously unnoticed and even counter-intuition observations, which may inspire future machine intelligence research.

To summarise, our contribution is three-fold: \textbf{(1)~Dataset}. We construct \TVD, a large-scale cinema/TV dramatic corpus that is segmented according to independent plots for the first time and has the most informative narrative data. It also introduces, to our knowledge, the first \textit{directly} public BSC collection (with 26 million BSCs inside).
Remarkably, \TVD is the first non-English multimodal corpus in the cinema/TV domain. \textbf{(2)~Model}. We released the first \textit{open-source} pre-trained model and weights for processing multimodal cinema/TV-drama data. It is also the first attempt of applying a unified network architecture to solve different tasks in this field. Built with state-of-the-art architectures, our method can serve as a strong baseline in follow-up research. \textbf{(3)~Tasks and Experiments}. Based on Cognitive Science research, we introduced three representative tasks and conducted extensive empirical tests. The novel insights reported (including counter-intuition findings) can benefit future multimodal studies.

\section{Related Work}
\label{sec:rw}

\noindent\textbf{Datasets.} Digital entertainment such as movies and TV dramas are attracting much research attention from the multimodal research field, as they offer a valuable snapshot of situations and scenarios in the real-world. 
LSMDC~16~\citep{lsmdc2015cvpr,lsmdc2017ijcv}, a combination of sub-datasets in several related domains, is the pioneer in this direction. 
Its success sparked the creation of a number of follow-ups: Pororo~\cite{pororo2017arxiv} which contains animation internal dialogues and external human summaries, YMS~\citep{yms2018cvpr} which comprises human-narrated movie plot summaries, and CMD~\citep{cmd2020accv} which pairs key clips with descriptions. Later resources including MovieNet~\citep{movienet2020eccv}, MAD~\citep{mad2022cvpr}, and SYMON~\citep{symon2022arxiv} not only expanded the scale of data and added more detailed descriptions (see statistics in Tab.~\ref{tab:compare}), but also introduced more diverse tasks, e.g., genre prediction, segment ordering, and multimodal retrieval. 

However, as discussed in~\cref{sec:introduction}, these corpora are limited in three major aspects (i.e., segmentation according to shots/scenes rather than a slot, lack of empathy and context in the narratives, and absence of long-form relationships due to short duration), constraining the scope and utility. The proposal of Movie101~\citep{yue2023movie101} gives a feasible solution, but its data cannot be obtained directly, and it lacks detailed plot descriptions. The \TVD dataset, in contrast, is constructed to mitigate these issues.

\noindent\textbf{Models.} Although recent years have witnessed many studies that conducted experiments using the above cinema/TV multimodal datasets~\citep{lsmdc2017ijcv,movienet2020eccv,symon2022arxiv,movie22022arxiv,yu2021transitional}, to the best of our knowledge, the model weights of \textit{neither} the methods proposed in these works \textit{nor} the baselines introduced in the original corpus papers are publicly available. Moreover, the architectures of these algorithms vary as they focus on different problems. As a result, it is difficult for follow-up works in this research vein (including ours) to recycle the models built in earlier studies, leading to extra training costs, unnecessary carbon footprint, as well as barriers to comparing different algorithms in a fair way.

To address this issue, for the first time, we align the level of machine learning tasks in the cinema/TV domain with the stage of human intelligence defined by Cognitive Scientists. Aiming to provide a strong and reusable baseline, we proposed a framework that can solve different tasks with a unified network design. We promise to release our model weights, so that follow-up works no longer need to develop their approaches from scratch.

\noindent\textbf{Bullet Screen Comments.} Bullet Screen Comments (BSCs) refer to the fashionable type of comments that fly across the screen like bullets when the video is playing. On the video timeline, a BSC appears at the point where it is initially posted by the user and hovers for a few seconds.
Compared with traditional static comments, BSCs are more likely to be strongly associated with specific dramatic pieces.
They make video viewers feel like having ``real-time'' discussions with others. Users can also make active interactions by upvoting BSCs they favour. 
See the appendix for a real-world screenshot.

A group of studies has validated the usefulness of BSCs in helping multimodal methods to model video semantics~\citep{bs32018movie,bs2www2020,bs12022aaai,yang2019time,wang2020videoic}. However, as far as we are aware, none of these works have the copyright of the BSCs, and there is no \textit{direct} download URLs have been provided. Besides, they have never investigated the role of BSCs in processing dramatic plots, either. In \TVD, we included more than 26 million BSC sentences, making the first public BSC dataset. Additionally, it is also the largest text corpus in the cinema/TV domain to date, whose scale is sufficient to bootstrap the training of large language models. Moreover, for the first time, we empirically justified that BSCs can effectively enhance systems' ability to modelling complex plots. 
\section{Corpus Construction}
\label{sec:construction}

\noindent\textbf{Overview.} Our data is obtained from online TV drama channel\footnote{\url{https://v.qq.com/channel/tv}.}, from which we downloaded the original videos, their well-structured metadata (abstract, release year, protagonist list, etc.), the corresponding BSCs, and plot descriptions composed by professional editors. 
Among the huge pool of available TV drama series, we manually picked out the high-quality ones that (1) have a view count of no less than 100,000, (2) contain more than one episode, and (3) each episode has received at least one BSC.

After human annotations, as shown in Tab.~\ref{tab:compare}, we ended up with 83 TV drama series, which consist of 1,106 episodes. Their total duration is 39,900 minutes, i.e., 36 minutes per episode.
Consistent with existing multimodal resources~\citep{movienet2020eccv,movie22022arxiv}, we respectively set the frame rate and video resolution at 480$\times$270 and 25 per second.
The encoding format is H264, and the encapsulation format is AVC1.
We harvested 24,875 clips, with an average duration of 63 seconds. We utilised FFmpeg to conduct uniform interceptions, yielding 4 screenshot frames per clip. 
In particular, to eliminate the influence of subtitles and video website watermark icons in downstream tasks, we mask them in screenshots separately.
We also collected Chinese subtitles within the clips using an open-source OCR toolkit, with an average of 9 words per subtitle.
Each clip corresponds to a standalone plot described in a sentence. The mean word count of such sentences is 40 words, leading to a total volume of 918,432 words for all clips.
As for BSCs, after applying an internal filter for content compliance and only preserving sentences that have a word count between 5 and 50, we finally included 26,438,007 samples (i.e., averagely 23,904 per episode and 663 per minute) into \TVD, with an average length of 13 words. 
We carefully aligned all these modalities with timestamps.

\label{sec:genre_tag_sys}



\begin{wrapfigure}{r}{0.6\textwidth}
\centering
    \includegraphics[width=0.55\textwidth]{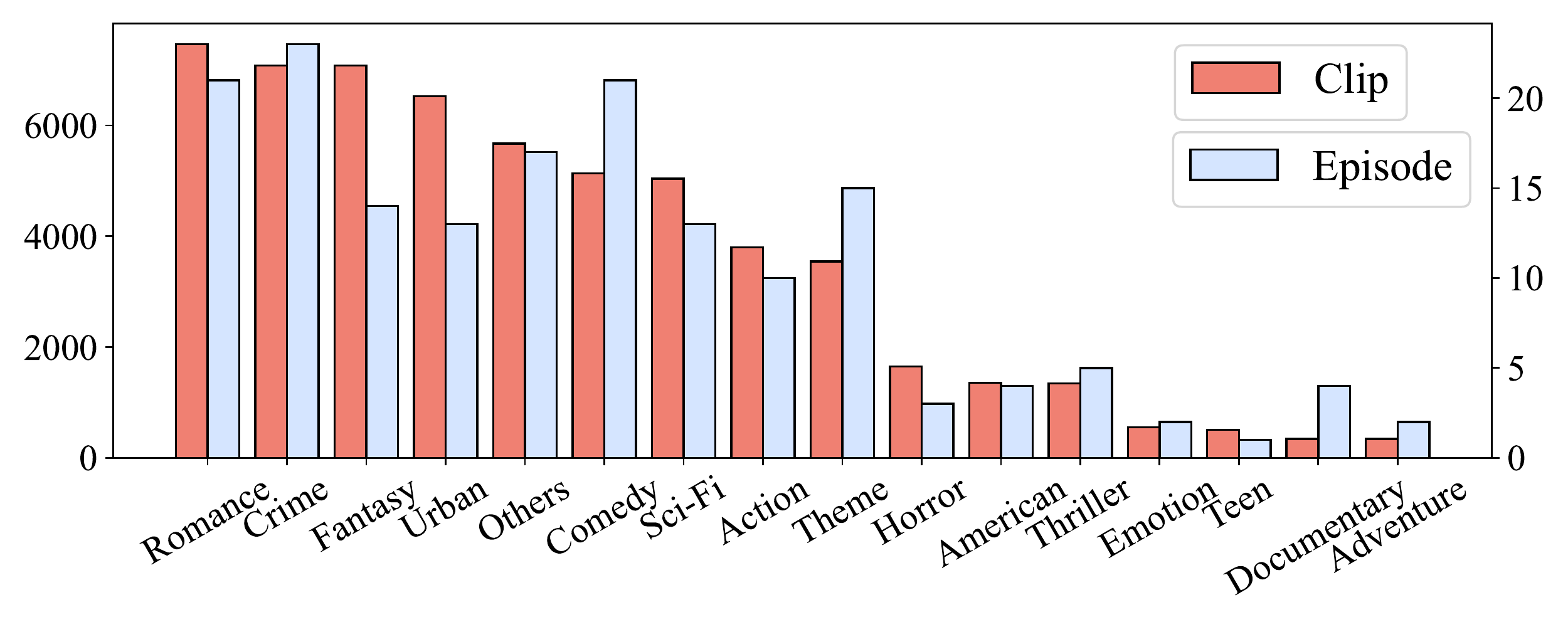}
    \vspace{-0.2in}
    \caption{
    Distribution of \TVD genre tags on the first layer.
    The \colorbox{myred}{left} and \colorbox{myblue}{right} axes are for the numbers of clips and episodes, respectively.
    }
    \label{fig:genre_tags}
    \vspace{-0.1in}
\end{wrapfigure}

\noindent\textbf{Genre Tag System.} To facilitate tasks such as genre classification (see~\cref{sec:tasks}), we supplement the metadata of each of the 83 TV dramas with fine-grained genre tags. Through literature review, we found that existing academic genre tag systems may be unsuitable for TV dramas. On the one hand, their granularity tends to be too coarse (e.g., \citep{simoes2016ijcnn}, \citep{moviescope2019arxiv}, and \citep{movienet2020eccv} respectively provide 4, 13, and 24 categories only) to accurately distinguish different dramatic types. On the other hand, most of these tag systems focus on movies, thus failing to cover several important TV drama genres (e.g. Unit Drama). Therefore, we introduce a two-level system with 58 genre tags, which is presented in the appendix. As this system has been applied to a major Chinese video website\footnote{\url{https://v.qq.com/}.} where it was verified by commercial successes (e.g., achieving more than 150 million Daily Active Users), we recommend it to the wider community of video/multimodal processing for research use.


\begin{wrapfigure}{r}{0.6\textwidth}
\centering
\subfigure{
\begin{minipage}[t]{0.33\linewidth}
\centering
\includegraphics[width=1.0in]{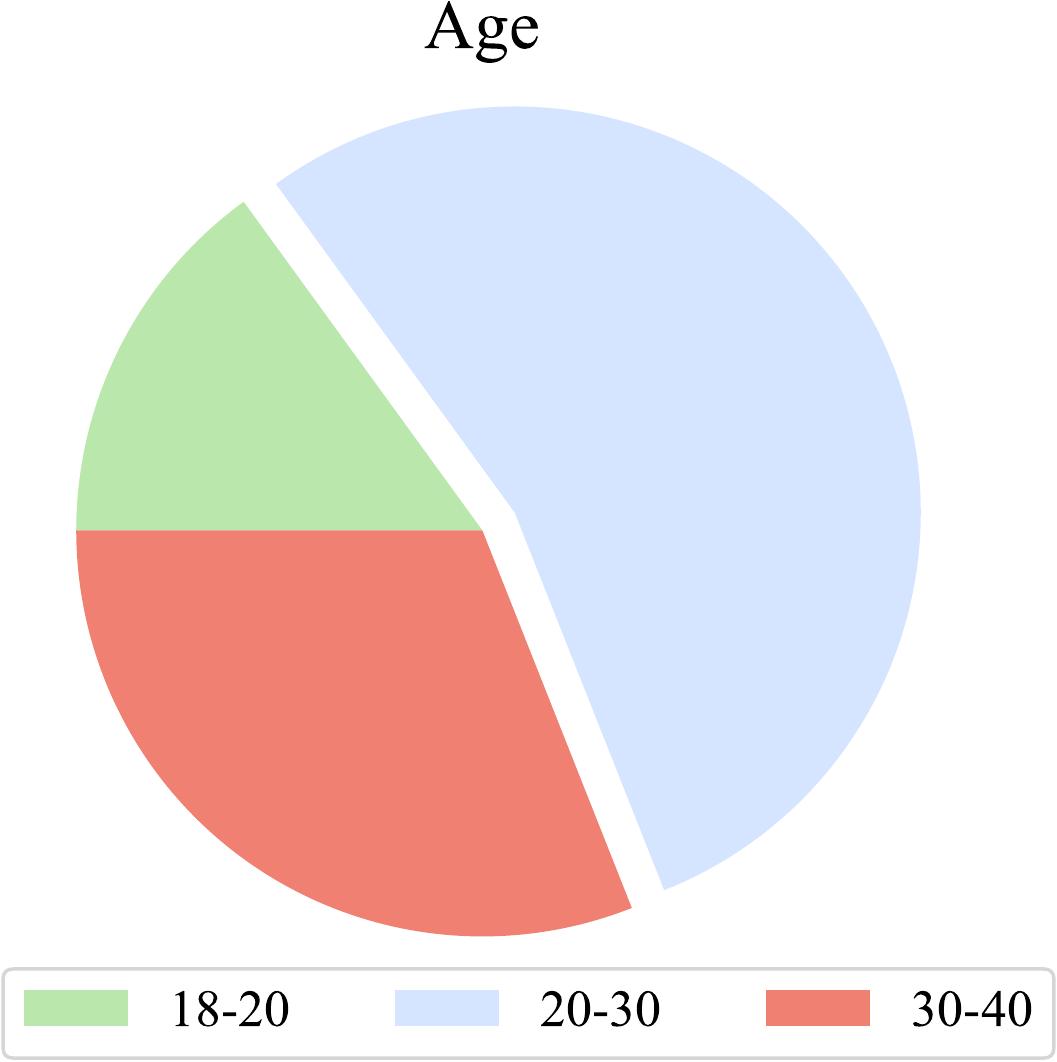}\\
\end{minipage}%
}%
\subfigure{
\begin{minipage}[t]{0.33\linewidth}
\centering
\includegraphics[width=1.0in]{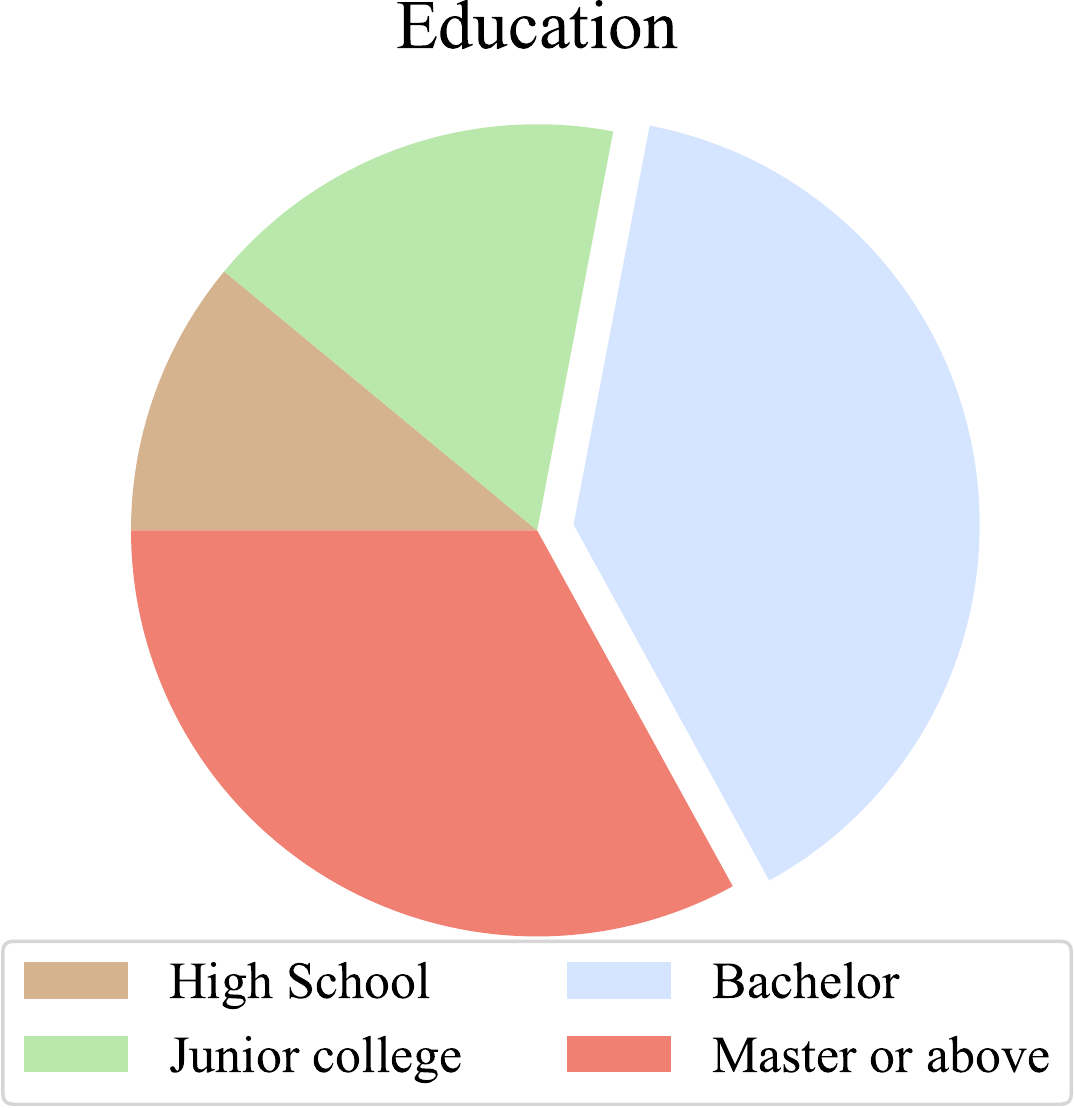}\\
\end{minipage}%
}%
\subfigure{
\begin{minipage}[t]{0.33\linewidth}
\centering
\includegraphics[width=1.0in]{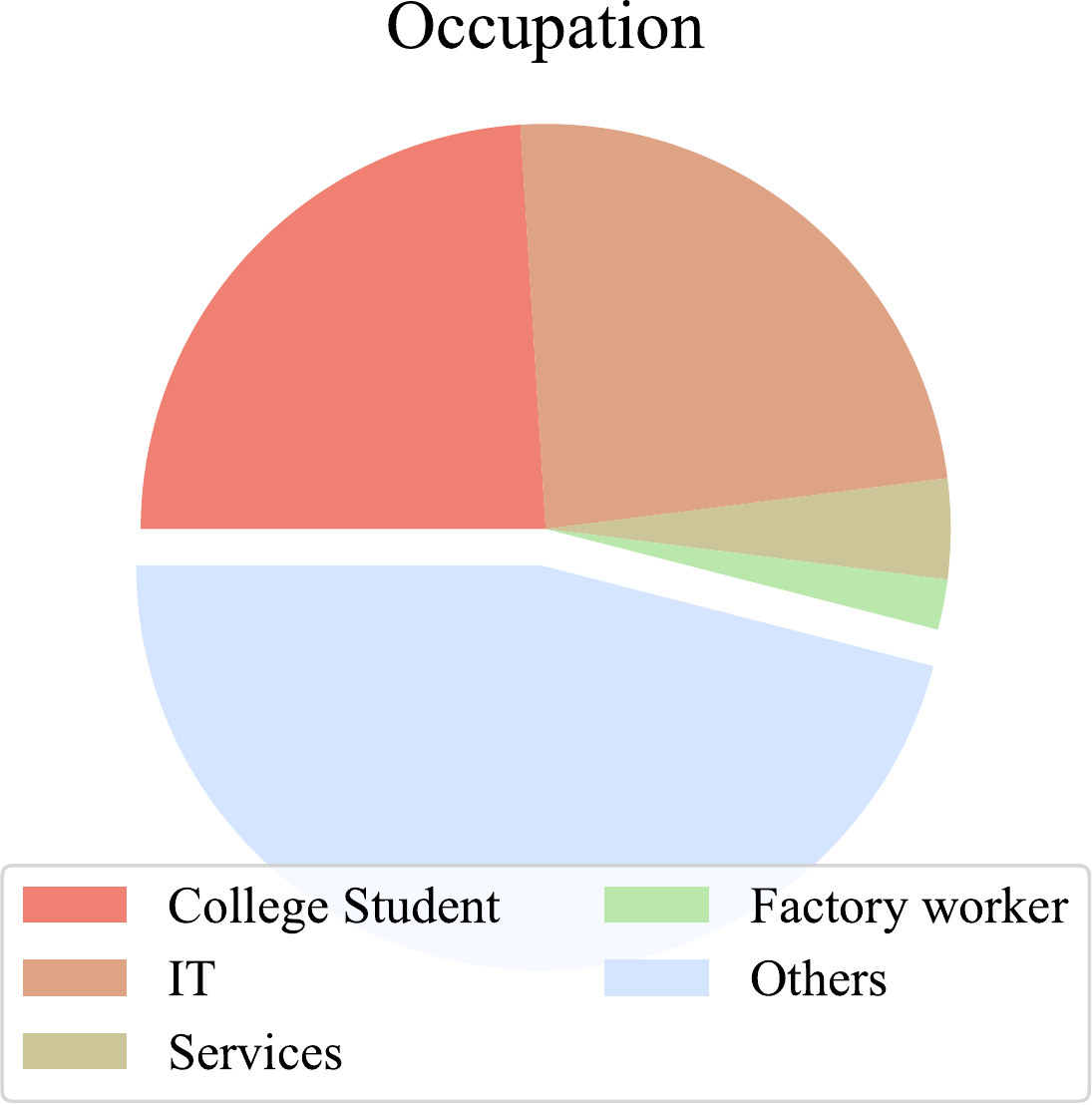}\\
\end{minipage}%
}%
\centering
\caption{Summary of annotator background.}
\label{fig:pie}
\vspace{-0.1in}
\end{wrapfigure}

\noindent\textbf{Annotating Plot-Oriented Clips.} We recruited Chinese native speakers from a crowd-sourcing platform\footnote{\url{https://soho.qq.com/home}}. 
Considering the cognitive level requirements due to the complexity of plot understanding~\citep{dev-theory-mind,child-theory-mind}, we only accepted annotators who have at least finished high school education. Eventually, 449 human annotators were selected. Their ages range from 18 to 40 years old. Almost half of them are college students or Internet practitioners (either accounting for 24.1\%). The entire annotation work cost 1,183 hours of human labour in total, i.e., averagely 2.37 hours for each annotator, 1.07 hours for each episode, and 2.87 minutes for segmenting each plot-oriented clip. See more statistics in Fig.~\ref{fig:pie}.

When segmenting one episode, an annotator received the complete video and the corresponding plot narrative which was written by professionals and crawled from the web (see above). Normally one such narrative has multiple sentences, with each defining a standalone plot. 
The annotator was first required to watch the episode in its entirety.  
Then, for each plot-focused sentence, the annotator was asked to choose from three options: `\textit{Video is not playable}', `\textit{Plot in this sentence is covered in one video span}', and `\textit{Plot in this sentence is not covered by one video span}' (the selection yielded an overall percentage of 0\%, 95.3\%, and 4.7\%).
We found that `\textit{Plot in this sentence is not covered by one video span}' was selected when the text seemed too vague or the annotators believed there are multiple plots in a single sentence.  In such a rare case, we made manual checks to decide whether to remove directly or divide the sentence into finer-grained pieces.
In most cases (i.e., `\textit{Plot in this sentence is covered in one video span}' was selected), the annotator was asked to mark the start and end timestamps of the video clip corresponding to the plot described in the sentence. During a post-processing step, we removed all the clips that are overly long (i.e., more than 10 minutes). 

To assess the annotation quality, we randomly selected 10\% samples and assigned them to two independent annotators.
We use Inter-Annotator Agreement~\citep{iaa1996pm} as the metric. Specifically, if the ratio of the intersection and union of the plot intervals given by different annotators is greater than 60\%, the annotations are considered to be consistent. 
Since the final Kappa score is 0.7152, which is within (0.6, 0.8], the result is validated as a "substantial agreement". This proves that the annotations of \TVD are of high quality.
\section{Cognitive-Inspired Tasks}
\label{sec:tasks}

To demonstrate the usefulness of \TVD in the development of algorithms for real-world applications, we conducted experiments on three representative tasks. Specially, we argue that they are closely related to the development milestones of human intelligence. To be exact, at the age of 3, children start to occupy the capability of discriminating the genres of TV content~\citep{wright1994young}. By 10 years old, as memory grows, children can follow major plot ingredients presented explicitly, thus being able to identify one cinema/TV modality (e.g., video) given another (e.g., text)~\citep{gunter2005children}. Writing empathic and context-aware plot descriptions, however, is a complicated skill which requires not only intelligence but also proper training, so most children have trouble mastering it until the end of the Formal Operational Stage (around 16 years old)~\citep{piaget2003psychology}. Analogously, by visiting the tasks of Genre Classification, Plot Retrieval, and Plot Text Generation, one can also study multimodal systems from a shallow level to a deep level, thus revealing the weaknesses and limitations of tested algorithms.

\noindent\textbf{Genre Classification.} Genre Classification is the task of labeling a sample with one or more genre tags, which is similar to the multi-classification problems concerned by previous works in the cinema domain~\citep{moviescope2019arxiv,movienet2020eccv}.
Since the plot already provides concentrated genre information (e.g., fighting plots in action dramas, dating plots in romance dramas), we take the plot-oriented clips as the primary unit to be processed.
As for the metric, following common practice in multi-classification, we calculated the F1 scores. Besides the Micro setup, considering the unbalanced genre distribution of clips (see Fig.~\ref{fig:genre_tags}), we also benchmark the Macro one.

\noindent\textbf{Plot Retrieval.} This task asks a model to retrieve matched results whose modality is different from that of the input~\citep{cmr2016arxiv}.
As all modalities of \TVD are made parallel, it serves as a good test bed for cross-modal retrieval.
Specifically, in this work, we experimented with two different retrieval settings: Image-Text, and Video-Text. Among them, the video mode contains both visual signals and text signals (i.e., image frames and subtitle texts).
For the text side, we used plot-focused sentences rather than the BSCs. The main reason is that a BSC can be quite general, e.g., ``\textit{Actor A is so beautiful!}'', in which case it may point to a huge number of correct video/image candidates, making the evaluation difficult. 
Additionally, we did not consider the retrieval between image and video as the former is directly captured from the latter.

We reported two groups of results for the Plot Retrieval task. The first is Hit@$n$ ($n \in \{1,5,10\}$), i.e., the fraction of correct answers that rank in the top $n$ retrieved samples.
The second is Matching Accuracy (following \citep{meter2022cvpr}). To be exact, we asked the model to traverse all possible pairings between samples from two modalities and calculated the overall accuracy.

\noindent\textbf{Plot Text Generation.} Beyond the above two understanding-only tasks, we added this more complex generation problem.
In the cinema/TV domain, previous studies have attempted to develop methods that can take video clips (including frames and subtitles) as input and produce scene descriptions as output, so as to help visually impaired individuals enjoy movies~\citep{vc12017itm,vc22020cvpr}. Our Plot Text Generation task is more challenging than that, as the tested model is additionally expected to perceive the dramatic context and the mental states of roles. To the best of our knowledge, this is one of the most ambitious multimodal text generation goals in the community~\citep{wang2021survey,zelaszczyk2023cross}.
In our study, we selected three popular types of automatic metrics: BLEU (@$\{1,2\}$)~\citep{bleu2002acl}, METEOR~\citep{meteor2005acl}, and CIDEr~\citep{cider2015cvpr}. 


\section{A Strong Baseline of PTVD Benchmark}

\subsection{Framework}
\label{sec:framework}

As discussed in \cref{sec:rw}, previous multimodal methods that process the cinema/TV domain are all task-dependent. Moreover, none of them have publicly shared model weights, let alone trained networks for Chinese text (NB: all past multimodal movie/TV-drama corpora are in English). Therefore, there does not exist a direct baseline that we can utilise for customisation or benchmark purposes.
To tackle this challenge, based on METER~\citep{meter2022cvpr}, a state-of-the-art architecture for multimodal learning, we developed a unified framework that can attack various cinema/TV processing tasks.

\begin{wrapfigure}{r}{0.6\textwidth}
\centering
    \includegraphics[width=0.45\textwidth]{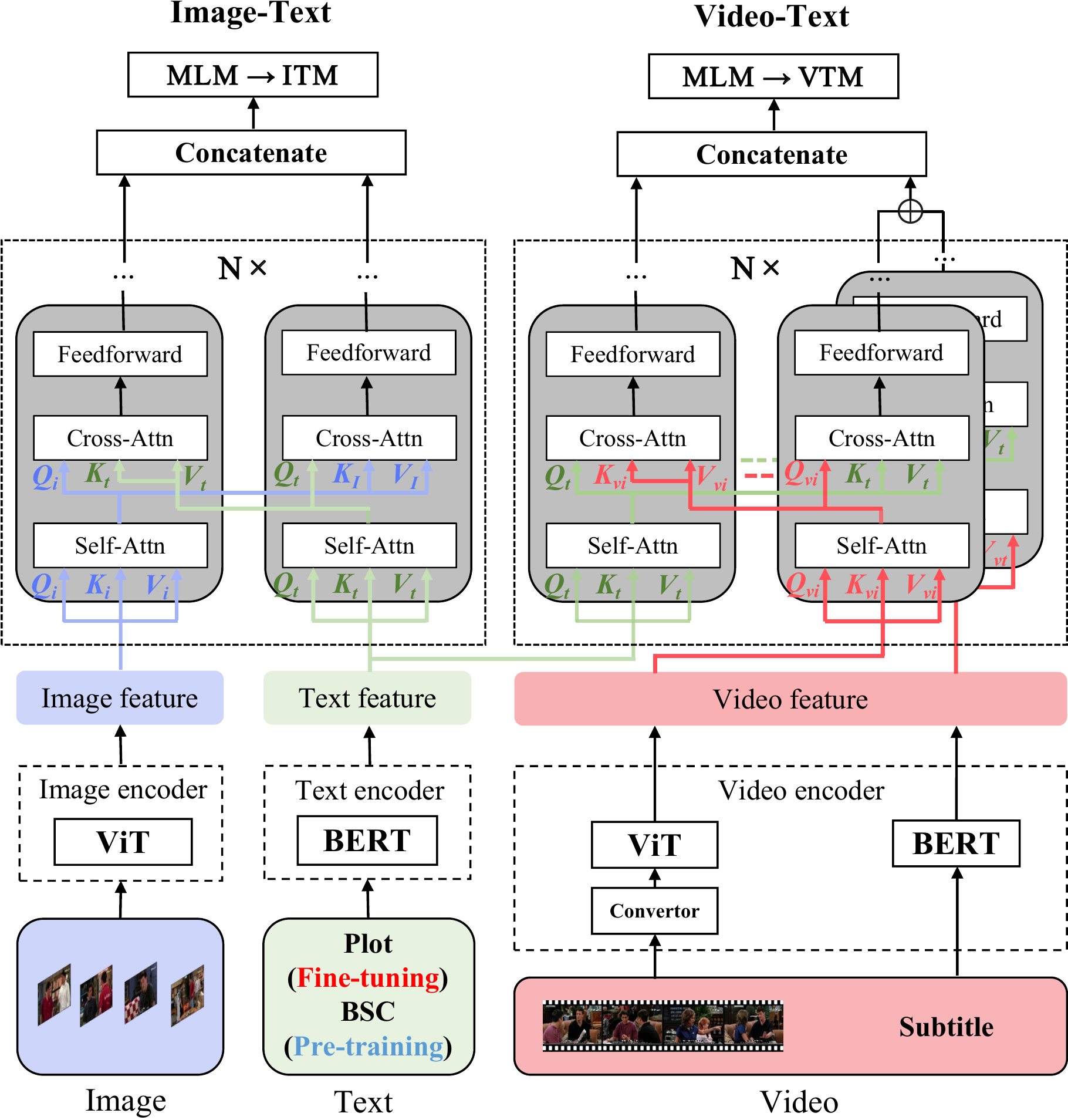}
    \caption{
    Architecture of the pre-training framework.
    }
    \label{fig:framework}
\end{wrapfigure}

\noindent\textbf{Network structure.} Shown in Fig.~\ref{fig:framework}, aiming to serve as a future baseline, our algorithm first adopts the most popular text encoder, BERT~\citep{bert2018arxiv}, for text input, and a very promising vision encoder, ViT~\citep{vit2020arxiv}, for image and video.  
The vanilla METER does not support video input, so we employ the approach of \citep{4f2021cvpr} and add a signal converter which uniformly samples four frames from each clip. In the Image-Text setting, the encoded features are fed into co-attention fusion modules to acquire a multimodal representation. In the Video-Text setting, the encoded video features, which include images and subtitles, are directed toward two independent attention fusion modules. A bitwise addition operation is then implemented before concatenating with the plot text to learn the multimodal representation.

\noindent\textbf{Pre-training with BSCs.} It is known that the perception and expression abilities of young children can see a development boost if they verbally interact with their parents while watching TV programmes~\citep{bullet-screen-pyschology}. Motivating by how BSCs are applied for interactions between video viewers (see \cref{sec:rw}), we proposed to conduct large-scale pre-training using the BSC collection of \TVD, which is the biggest text corpus in the cinema/TV domain.
To be specific, we first perform Masked Language Modelling (MLM)~\citep{bert2018arxiv} with BSCs. Next, given the timestamp of each BSC, we captured one screenshot frame. We also clipped a 60-second video piece that surrounds the timestamp. In this way, we obtained more than 26 million BSC-image and BSC-video pairs, respectively. We made use of these two groups for Image-Text Matching (ITM) and Video-Text Matching (VTM)~\citep{itm2019nips}. Note that Image-Text and Video-Text sides are independent, and only parameters of the co-attention fusion modules are updated. 
See the appendix for comprehensive hyper-parameters.

\noindent\textbf{Fine-tuning Scheme.} For Genre Classification, we input all possible pairings of the involved modalities. Specially, as a BSC may only correspond to the content of a short video span, for each clip we fed the concatenation of the top 10 most upvoted BSCs to the model.
All output embeddings are then concatenated and passed to a Multi-Layer Perceptron (MLP), which is optimised against the cross-entropy loss. Due to the multi-label nature of the task, for each tag we independently trained one binary classifier.

As for Plot Retrieval, on the Image-Text side of our framework, we performed Img2Txt and Txt2Img retrievals. The Video-Text side, correspondingly, is trained on Vid2Txt and Txt2Vid retrievals.
We trained a MLP head to predict if the inputs from two modalities make a correct match.
The prediction score is utilised directly to obtain Hit@$n$. Its binarised version (we set the threshold at 0.5) is used to calculate Matching Accuracy. 


While the original METER does not support text generation, we ``made it speak'' through the workaround of \citep{captioning2020eccv}. Specially, we introduced causal masking to the attention units within the co-attention fusion module of the video-text side, which turned on the auto-regressive inference mode of our framework. We argue that this strategy can be applied to other text generation tasks as well. 
See more implementation details in the appendix.

\subsection{Experiment Results}
\label{sec:experiment_res}

\noindent\textbf{Genre Classification.} From Tab.~\ref{tab:ablation_genre_cls}, we confirm that multimodal data does help in the Genre Classification task, as the configuration where all modalities are involved clearly outperformed the ablation setups. 
Moreover, we noticed that text signals, especially the plot text, substantially contribute more than visual ones. Such a trend is also observed in uni-modal experiments, which are presented in the appendix due to the length limit.

\begin{wraptable}{r}{0.5\textwidth}
\centering
\scriptsize
\begin{tabular}{lcccc}
\toprule
\textbf{Modality} & \textbf{Mirco-F1} & \textbf{Drop} & \textbf{Marco-F1} & \textbf{Drop} \\
\midrule
All & 95.3 & - & 82.8 & - \\
\textit{w/o} Plot & 78.7 & 16.6 & 75.4 & 7.4 \\
\textit{w/o} BSC & 90.8 & \textcolor{white}{0}4.5 & 80.2 & 2.6 \\
\textit{w/o} Video & 90.9 & \textcolor{white}{0}4.4 & 80.0 & 2.8 \\
\textit{w/o} Image & 92.5 & \textcolor{white}{0}2.8 & 80.8 & 2.0 \\
\bottomrule
\end{tabular}
\caption{Results of Genre Classification. We list the relative F1 drop in each ablation setup.}
\label{tab:ablation_genre_cls}
\end{wraptable}


One interesting observation is that, while including more modalities substantially improves the classification performance (both Micro and Macro), meanwhile it consistently aggregates the gap between Micro and Macro scores. For instance, without plot text, Micro F1 is higher than Macro F1 by 3.3, yet using plot text widens this difference to 12.5. Larger Micro-Macro gaps indicate a more significant imbalance between the model performance in various categories. Further investigations revealed that, frequent genres (e.g., \textit{Fantasy Romance}, \textit{High-IQ Crime}, and \textit{Situation Comedy}) which are tagged on many episodes (the model thus performs strongly even with fewer modalities) tend to benefit more than the less popular ones (e.g., \textit{Documentary and Adventure}). Therefore, we conclude that the improving effects of leveraging modalities can be biased towards frequent tags, which, to our knowledge, is the first of such findings in the field of cinema/TV multimodal learning.


\begin{table*}[t]
\centering
\scriptsize
\begin{tabular}{ccc|ccc|ccc|c}
\toprule
\multirow{2}{*}{\textbf{Modalities}} & \multirow{2}{*}{\textbf{Pre-Train}} & \multirow{2}{*}{\textbf{Fine-Tune}} & \multicolumn{3}{c|}{\textbf{From Text}} & \multicolumn{3}{c|}{\textbf{To Text}} & \multirow{2}{*}{\begin{tabular}[x]{@{}c@{}}\textbf{Matching}\\\textbf{Accuracy}\end{tabular}} \\
\cmidrule{4-9}
 &  &  & \textbf{Hit@1} & \textbf{Hit@5} & \textbf{Hit@10} & \textbf{Hit@1} & \textbf{Hit@5} & \textbf{Hit@10} &  \\
\midrule
\multirow{3}{*}{Image-Text} & \checkmark & \ding{55} & 0.3 & 2.2 & \textcolor{white}{0}4.0 & 0.3 & 1.9 & \textcolor{white}{0}4.5 & 50.3  \\
 & \ding{55} & \checkmark & 0.4 & 2.8 & \textcolor{white}{0}6.1 & 0.5 & 3.1 & \textcolor{white}{0}5.9 & 59.1 \\
 & \checkmark & \checkmark & 0.5 & 3.4 & \textcolor{white}{0}7.0 & 0.7 & 4.8 & \textcolor{white}{0}7.5 & 63.6 \\
\midrule
\multirow{3}{*}{Video-Text} & \checkmark & \ding{55} & 0.2 & 0.6 & \textcolor{white}{0}1.9 & 0.1 & \textcolor{white}{0}0.6 & \textcolor{white}{0}1.4 & 79.2 \\
 & \ding{55} & \checkmark & 1.0 & 6.0 & 10.9 & 1.3 & \textcolor{white}{0}6.4 & 12.1 & 81.5 \\
 & \checkmark & \checkmark & 2.2 & 9.6 & 17.5 & 2.4 & 10.7 & 18.5 & 89.1 \\
\bottomrule
\end{tabular}
\caption{\label{tab:retrieval}
Results (\%) of Cross-Modal Retrieval. 
}
\vspace{-0.1in}
\end{table*}

\noindent\textbf{Plot Retrieval.} As shown in Tab.~\ref{tab:retrieval}, on the one hand, in all setups tested, fine-tuning with plot text substantially improves Plot Retrieval performance. BSCs (used at the pre-training stage) also benefit the final result (which is expected in \cref{sec:framework}), although not as significantly as the professionally written plot text. 

On the other hand, despite being quite effective on the cognitively shallower Genre Classification task, our large-scale framework performs unsatisfactorily on more challenging Plot Retrieval tasks. To be exact, in all setups, even Hit@10 scores can only reach up to 18.5\%. As for the Matching Accuracy, considering a random guess will yield 50\%, the results, especially those of the Image-Text group, are to be improved. In particular, we notice that when setting text (i.e., plot-focused sentences) as the retrieval target, the model performance is significantly ($p<0.5$) better than configuring videos or images as the target. This suggests that the current framework tends to distinguish \TVD's text signals better than the visual ones.


On conventional scene-oriented corpora, previous studies reported that inputting video or key frame does not have a remarkable performance impact~\citep{keyframe2022revisiting}. However, this does not apply to plot-oriented \TVD, where the model employing video is consistently superior to the one using the image. At the same time, we have also reached the same conclusion as~\citep{lei2020tvr}, which suggests that video data comprising subtitles and images perform better in terms of retrieval performance compared to video data comprising only multiple images. For further details, please refer to the appendix. Besides, we found that improving the interpolation frequency, which consequently helps feed more information to the model, can further improve the retrieval scores  (see experiment results in the appendix). In summary, plot-oriented \TVD can access the model's ability to capture long-form relationships while previous scene-oriented datasets cannot.



\begin{wraptable}{r}{0.6\textwidth}
\centering
\scriptsize
\begin{tabular}{ccccccc}
\toprule
 & \textbf{Ref \#} & \textbf{P-T} & \textbf{BELU@1} & \textbf{BELU@2} & \textbf{METEOR} & \textbf{CIDEr} \\
\midrule
\multirow{2}{*}{\rotatebox{90}{Plot}} & \multirow{2}{*}1 & \checkmark & 18.2 & \textcolor{white}{0}9.6 & 8.0 & \textcolor{white}{0}9.0 \\
& & \ding{55} & 15.2 & \textcolor{white}{0}7.6 & 7.6 & \textcolor{white}{0}6.7 \\
\midrule
\multirow{2}{*}{\rotatebox{90}{BSC}} & \multirow{2}{*}1 & \checkmark& \textcolor{white}{0}8.6 & \textcolor{white}{0}3.9 & 3.7 & \textcolor{white}{0}8.0 \\
& & \ding{55}  & \textcolor{white}{0}9.9 & \textcolor{white}{0}4.1 & 4.7 & \textcolor{white}{0}9.1 \\
\midrule
\multirow{2}{*}{\rotatebox{90}{BSC}}& \multirow{2}{*}{5}  & \checkmark & 25.2 & 12.7 & 8.1 & \textcolor{white}{0}6.4 \\
& & \ding{55}  & 38.9 & 18.0 & 9.2 & \textcolor{white}{0}5.6 \\
\bottomrule
\end{tabular}
\caption{Results of Plot Text Generation. ``Ref \#'' and ``P-T'' respectively represent the number of golden references and whether pre-training on BSCs is enabled.} 
\label{tab:plot_main_tab}
\end{wraptable}

\noindent\textbf{Plot Text Generation.} The results in Tab.~\ref{tab:plot_main_tab} are partial counter-intuition: on the one hand, pre-training with BSCs benefits the model on Plot Retrieval, and
constantly has a positive impact on Plot Text Generation. On the other hand, surprisingly, pre-training on BSCs brings a negative impact on all automatic evaluation scores in BSC generation. 
Even if we expanded the golden reference set to the top 5 most upvoted BSCs, the gap (BELU\@1, 2) brought by pre-training become even greater.
We have reason to suspect that the inconsistent distribution between BSC and plot text is a potential cause. We train the framework to generate BSC, rather than using the BSC with the most votes in each clip as a golden reference. We left investigation into this matter as an important future work.

Moreover, eight different text entropy calculators~\citep{entropy2022acl} suggested that the distribution of plot text is more complex than BSCs (see statistics in the appendix). Considering BSCs are also more than plot text in terms of the total number, writing BSCs is intuitively less challenging than producing text plots. Nonetheless, empirically the generation scores of plot text are higher than those of BSCs (with one reference). In the future, we will dive deeper into \TVD for explanations. 


\setlength\tabcolsep{1.5mm}
\begin{table*}[t]
\centering
\tiny
\begin{tabular}{cll}
\toprule
\textbf{Video} & \multicolumn{2}{c}{\textbf{Text}} \\
\midrule
\multirow{4}{*}{\includegraphics[width=0.35\linewidth]{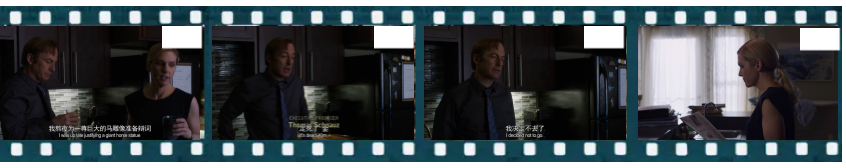}} & \multirow{2}{*}{Ground truth:} & \begin{CJK*}{UTF8}{gbsn}见吉米决定不去，她也就不再坚持。\end{CJK*} \\
 & & When Jimmy decided not to go, she stopped insisting. \\
 & \multirow{2}{*}{Generation:} & \begin{CJK*}{UTF8}{gbsn}吉米兴致不高，金没有什么反应。\end{CJK*} \\
 & & Jimmy was not very interested and Kim did not react. \\
 & & \\
\multirow{4}{*}{\includegraphics[width=0.35\linewidth]{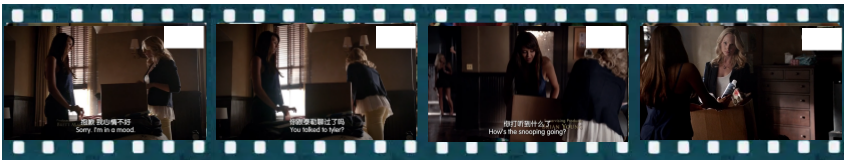}} & \multirow{2}{*}{Ground truth:} & \begin{CJK*}{UTF8}{gbsn}泰勒一直都没有和凯若琳联系，这让凯若琳有些生气。\end{CJK*}  \\
 & & Taylor has not been in touch with Caroline, which makes Caroline a little angry. \\
 & \multirow{2}{*}{Generation:} & \begin{CJK*}{UTF8}{gbsn}凯若琳很不高兴，她告诉艾琳娜自己还是要去找自己的母亲。\end{CJK*} \\
 & & Caroline is very unhappy, she tells Elena that she is still going to find her mother. \\
 & & \\
\multirow{4}{*}{\includegraphics[width=0.35\linewidth]{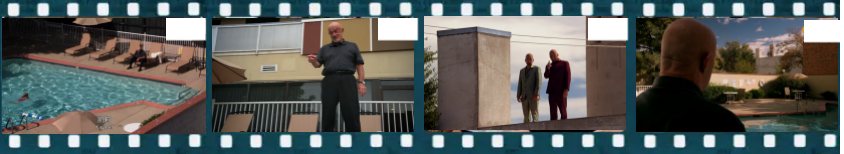}} & \multirow{2}{*}{Ground truth:} & \begin{CJK*}{UTF8}{gbsn}老迈克看到那两人将手做出手枪的模样指向凯莉。\end{CJK*}  \\
 & & Old Mike saw the two men point their hands like pistols at Kelly. \\
 & \multirow{2}{*}{Generation:} & \begin{CJK*}{UTF8}{gbsn}那两个手下还有枪，并指向了他。\end{CJK*} \\
 & & The two men still had guns and pointed them at him. \\
\bottomrule
\end{tabular}
\caption{\label{tab:plot_generation_case}
Examples of generated plot text.
}
\vspace{-0.2in}
\end{table*}
 
Lastly, we listed some Plot Text Generation examples in Tab.~\ref{tab:plot_generation_case} (with more in the appendix).
These cases reveal that, trained on \TVD, our framework is able to integrate situational context and mental states to some degree. 
However, it also makes mistakes, e.g., mistaking one role for another,  misunderstanding the relationships between roles, or misinterpreting the intentions of certain actions. 
In summary, using \TVD we show there is still much room for improvement in multi-modal learning on cinema/TV data.
\section{Conclusion}
\label{sec:conclusion}
We built \TVD, the first plot-oriented multi-modal dataset in the cinema/TV domain, which contains video clips that cover long-form relationships, informative plot descriptions written by professional editors, and the first Bullet Screen Comment collection that powers large-scale pre-training.
To provide a strong baseline for follow-up works, we developed the first framework that can handle various multi-modal problems with movie/TV-drama data, in a unified fashion.
With \TVD and our framework, we attacked three representative tasks which are motivated by Cognitive Science findings.
Our empirical observations (including counter-intuition ones) not only provides insights to future research, but also highlights the value of \TVD.


\bibliography{main}
\bibliographystyle{plain}

\appendix

\section{Dataset Details}
\label{sec:app_dataset_details}

We will provide more data statistics, detailed analysis, and sample display of \TVD.

\subsection{Genre Tag System}
\label{sec:app_genre_tag_system}

\begin{table*}[h]
    \centering
    \scriptsize
    \begin{tabularx}{\linewidth}{l|X}
    \toprule
    \makecell[c]{1st layer tag} & \makecell[c]{2nd layer tags} \\
    \midrule
    \begin{CJK*}{UTF8}{gbsn}奇幻\end{CJK*}/fantasy & \begin{CJK*}{UTF8}{gbsn}奇幻魔法\end{CJK*}/fantasy magic, \begin{CJK*}{UTF8}{gbsn}奇幻爱情\end{CJK*}/fantasy romance, \begin{CJK*}{UTF8}{gbsn}都市奇幻\end{CJK*}/urban fantasy \\
    \midrule
    \begin{CJK*}{UTF8}{gbsn}都市\end{CJK*}/urban & \begin{CJK*}{UTF8}{gbsn}都市奇幻\end{CJK*}/urban fantasy, \begin{CJK*}{UTF8}{gbsn}都市喜剧\end{CJK*}/urban comedy \\
    \midrule
    \begin{CJK*}{UTF8}{gbsn}喜剧\end{CJK*}/comedy & \begin{CJK*}{UTF8}{gbsn}都市喜剧\end{CJK*}/urban comedy, \begin{CJK*}{UTF8}{gbsn}恶搞闹剧\end{CJK*}/spoof farce, \begin{CJK*}{UTF8}{gbsn}黑色喜剧\end{CJK*}/dark comedy, \begin{CJK*}{UTF8}{gbsn}爱情喜剧\end{CJK*}/romantic comedy, \begin{CJK*}{UTF8}{gbsn}情景喜剧\end{CJK*}/situation comedy, \begin{CJK*}{UTF8}{gbsn}家庭喜剧\end{CJK*}/family comedy, \begin{CJK*}{UTF8}{gbsn}生活喜剧\end{CJK*}/life comedy, \begin{CJK*}{UTF8}{gbsn}搞笑\end{CJK*}/funny, \begin{CJK*}{UTF8}{gbsn}趣味\end{CJK*}/spice \\
    \midrule
    \begin{CJK*}{UTF8}{gbsn}惊悚\end{CJK*}/thriller & \begin{CJK*}{UTF8}{gbsn}悬疑\end{CJK*}/suspense, \begin{CJK*}{UTF8}{gbsn}惊悚\end{CJK*}/thriller \\
    \midrule
    \begin{CJK*}{UTF8}{gbsn}犯罪\end{CJK*}/crime & \begin{CJK*}{UTF8}{gbsn}高智商犯罪\end{CJK*}/high-IQ crime, \begin{CJK*}{UTF8}{gbsn}警匪犯罪\end{CJK*}/police and bandits crime, \begin{CJK*}{UTF8}{gbsn}烧脑罪案\end{CJK*}/brain burning crime, \begin{CJK*}{UTF8}{gbsn}侦探片\end{CJK*}/detective \\
    \midrule
    \begin{CJK*}{UTF8}{gbsn}科幻\end{CJK*}/sci-fi & \begin{CJK*}{UTF8}{gbsn}科幻动作\end{CJK*}/sci-fi action, \begin{CJK*}{UTF8}{gbsn}软科幻\end{CJK*}/soft sci-fi, \begin{CJK*}{UTF8}{gbsn}超自然\end{CJK*}/supernatural \\
    \midrule
    \begin{CJK*}{UTF8}{gbsn}动作\end{CJK*}/action & \begin{CJK*}{UTF8}{gbsn}科幻动作\end{CJK*}/sci-fi action, \\
    \midrule
    \begin{CJK*}{UTF8}{gbsn}爱情\end{CJK*}/romance & \begin{CJK*}{UTF8}{gbsn}奇幻爱情\end{CJK*}/fantasy romance, \begin{CJK*}{UTF8}{gbsn}熟龄爱情\end{CJK*}/mature romance, \begin{CJK*}{UTF8}{gbsn}曲折爱情\end{CJK*}/tortuous romance, \begin{CJK*}{UTF8}{gbsn}浪漫爱情\end{CJK*}/romantic love, \begin{CJK*}{UTF8}{gbsn}多角恋\end{CJK*}/polygonal love \\
    \midrule
    \begin{CJK*}{UTF8}{gbsn}题材\end{CJK*}/theme & \begin{CJK*}{UTF8}{gbsn}女性题材\end{CJK*}/female subject, \begin{CJK*}{UTF8}{gbsn}金融题材\end{CJK*}/finance, \begin{CJK*}{UTF8}{gbsn}律政题材\end{CJK*}/legal, \begin{CJK*}{UTF8}{gbsn}剧情故事\end{CJK*}/plot story, \begin{CJK*}{UTF8}{gbsn}行业剧\end{CJK*}/family comedy, \begin{CJK*}{UTF8}{gbsn}职场剧\end{CJK*}/workplace, \begin{CJK*}{UTF8}{gbsn}广告片\end{CJK*}/advertising \\
    \midrule
    \begin{CJK*}{UTF8}{gbsn}青春\end{CJK*}/teen & \begin{CJK*}{UTF8}{gbsn}青春\end{CJK*}/teen, \begin{CJK*}{UTF8}{gbsn}校园\end{CJK*}/school \\
    \midrule
    \begin{CJK*}{UTF8}{gbsn}情感\end{CJK*}/emotion & \begin{CJK*}{UTF8}{gbsn}父爱\end{CJK*}/paternal love, \begin{CJK*}{UTF8}{gbsn}婚姻生活\end{CJK*}/matrimony, \begin{CJK*}{UTF8}{gbsn}亲情\end{CJK*}/family affection \\
    \midrule
    \begin{CJK*}{UTF8}{gbsn}美剧\end{CJK*}/american & \begin{CJK*}{UTF8}{gbsn}美剧\end{CJK*}/american, \begin{CJK*}{UTF8}{gbsn}超级英雄\end{CJK*}/superhero \\
    \midrule
    \begin{CJK*}{UTF8}{gbsn}恐怖\end{CJK*}/horror & \begin{CJK*}{UTF8}{gbsn}恐怖\end{CJK*}/horror \\
    \midrule
    \begin{CJK*}{UTF8}{gbsn}冒险\end{CJK*}/adventure & \begin{CJK*}{UTF8}{gbsn}冒险\end{CJK*}/adventure \\
    \midrule
    \begin{CJK*}{UTF8}{gbsn}纪录片\end{CJK*}/documentary & \begin{CJK*}{UTF8}{gbsn}伪纪录片\end{CJK*}/mockumentary, \begin{CJK*}{UTF8}{gbsn}野史传奇\end{CJK*}/wild history, \begin{CJK*}{UTF8}{gbsn}人物传记\end{CJK*}/biography, \begin{CJK*}{UTF8}{gbsn}人物故事\end{CJK*}/character story \\
    \midrule
    \begin{CJK*}{UTF8}{gbsn}其他\end{CJK*}/other & \begin{CJK*}{UTF8}{gbsn}单元剧\end{CJK*}/unit play, \begin{CJK*}{UTF8}{gbsn}漫画改编\end{CJK*}/comic adaptation \\
    \bottomrule
    \end{tabularx}
    \caption{\label{tab:genre_tags}
    The hierarchical genre tags in \TVD.
    }
\end{table*}

Tab.~\ref{tab:genre_tags} lists the two layers of genre tags respectively in \TVD.
It can be found that viewers of TV drama source websites pay more attention to TV dramas with genre tags such as romance, crime, and fantasy.
At the same time, due to viewer preference, copyright purchase, and some hidden factors, some genre tags (e.g., documentary and adventure) are less, resulting in the imbalance of genre tags in \TVD.
Furthermore, when jointly considering the statistical results by genre tags of plots and TV dramas, it can be found that the proportion of plots in sci-fi and urban TV dramas is large, while the proportion of plots in comedy and theme TV dramas is small.
It shows that the plot density of TV dramas with different genre tags exists a gap, which is usually determined by the shooting methods and prominent objectives.

\subsection{Plot Clip}
\label{sec:app_plot_clip}

\begin{figure}[ht]
\centering
\includegraphics[width=1.0\textwidth]{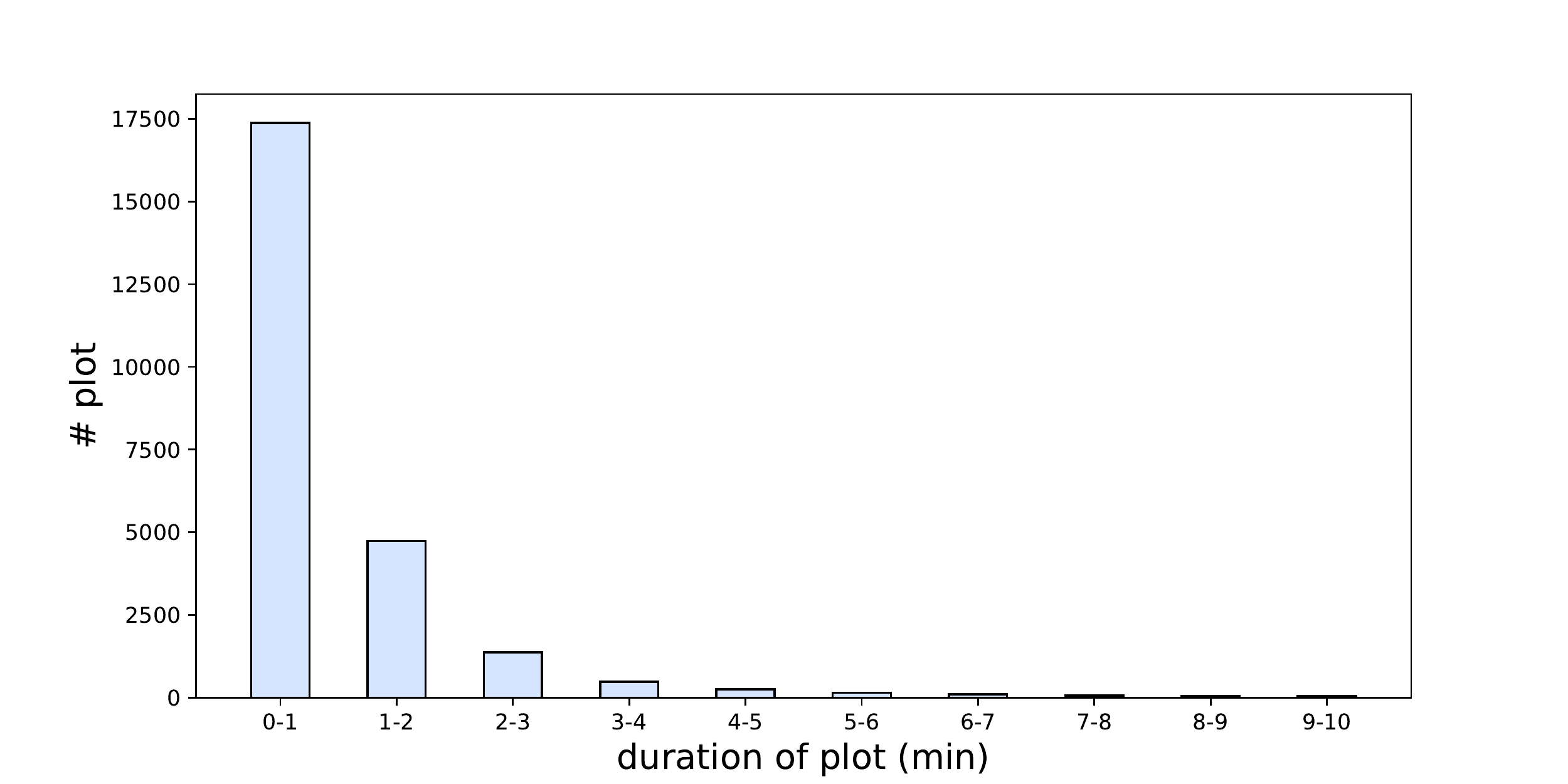}
\caption{
The distribution of duration of plot clip in \TVD.
}
\label{fig:duration}
\end{figure}

In \TVD, the core of parallel multi-modal data is the plot.
In Tab.~\ref{fig:duration}, we have counted the video clip duration in \TVD.
It shows the duration of the plot clip is mainly distributed between 0-2 minutes, which is significantly longer than existing datasets.
In addition, the plot clip can not only cover more long-distance dependent information, but also ensure the integrity of the plot.

\subsection{Bullet Screen Comments}
\label{sec:app_bsc}

\begin{figure}[ht]
\centering
\includegraphics[width=1.0\textwidth]{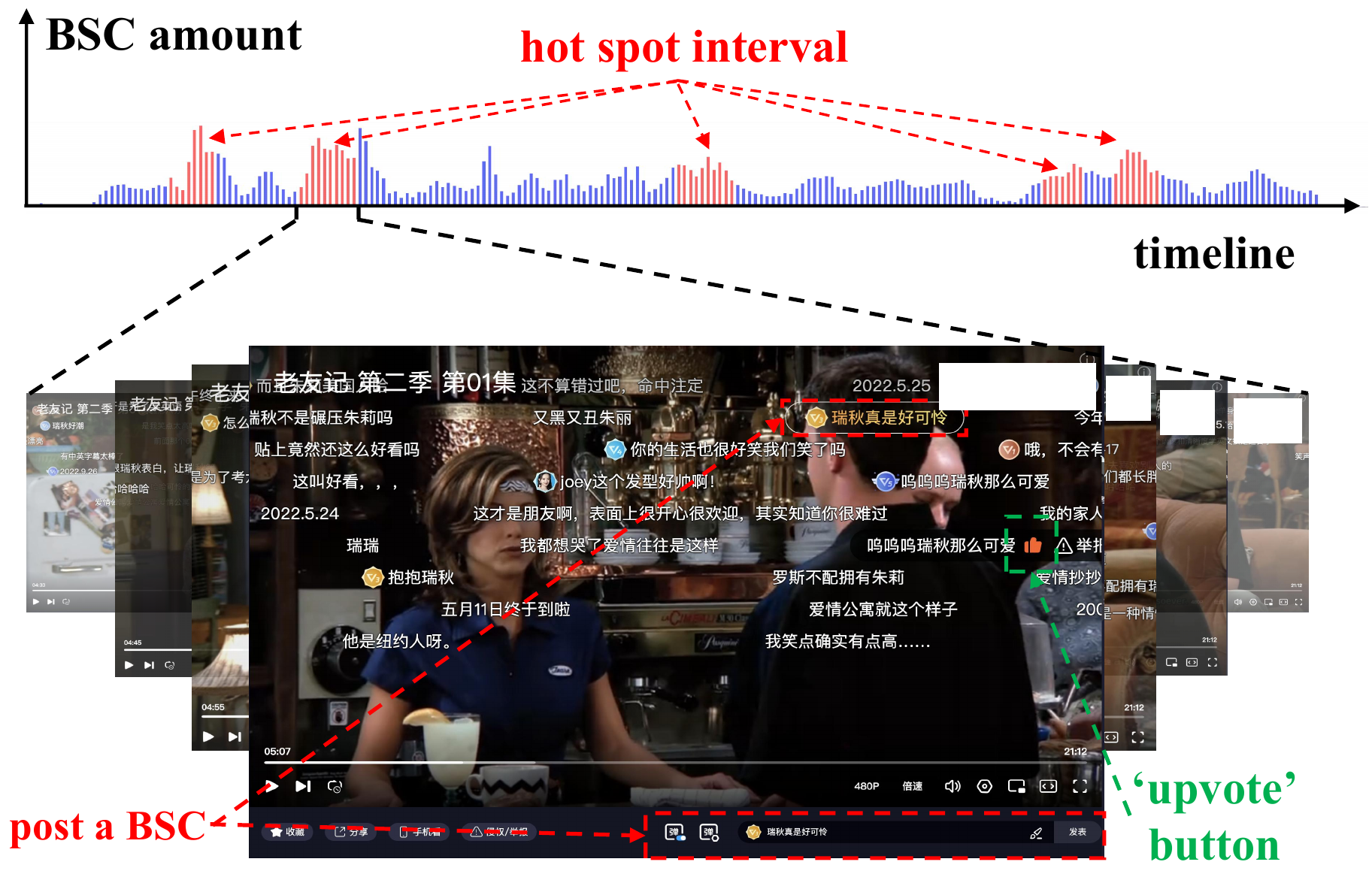}
\caption{
An example of BSC of the online video website.
}
\label{fig:bsc_example}
\end{figure}

As shown in Fig.~\ref{fig:bsc_example}, the BSC, as a real-time comment generated by viewers when watching a movie and TV dramas, will comment on the video content from various perspectives.
In particular, viewers will form a single round of opinion discussion with other viewers across time and space while watching TV dramas.
In addition, viewers express their agreement with the target BSC by clicking the `likes' button, which is also a simplified exchange of viewer opinions.
Therefore, the BSC data often contain a large amount of valuable behavior data, which can be used in commercial applications such as hot spot detection, topic mining, and movie review generation.

\subsubsection{Data statistics}
\begin{figure}[ht]
\centering
\includegraphics[width=0.7\textwidth]{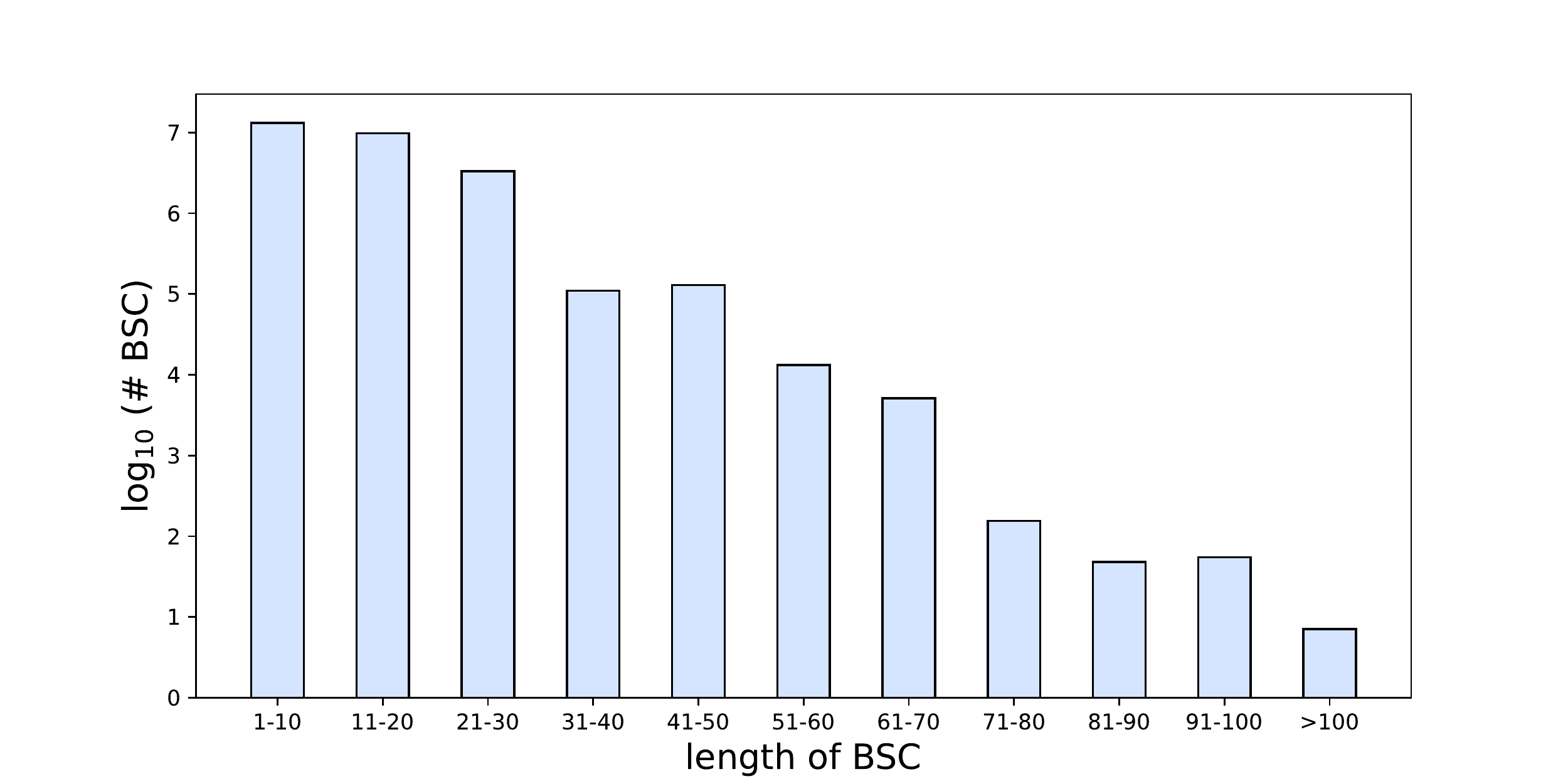}
\caption{
The length of the BSCs in \TVD.
}
\label{fig:bs_length}
\end{figure}

\begin{figure}[t]
\centering
\includegraphics[width=0.7\textwidth]{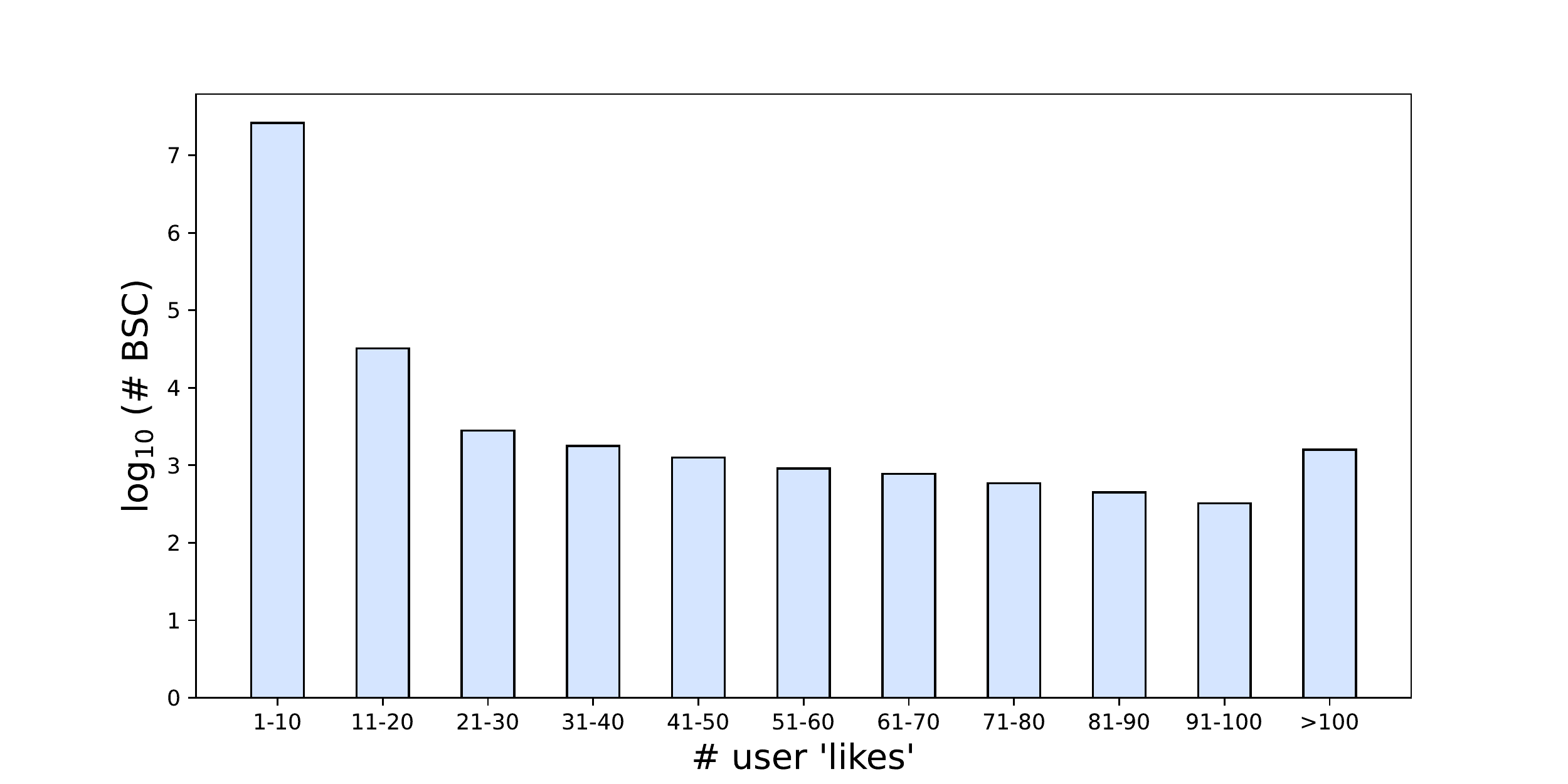}
\caption{
The distribution of user `likes' of the BSC in \TVD.
}
\label{fig:likes}
\end{figure}

In Fig.~\ref{fig:bs_length} and Fig.~\ref{fig:likes}, we have made detailed statistics on the information of the BSCs, including the length of the BSCs, the number of 'likes', and so on.
Among them, by analyzing the distribution of the length of the BSC, we can find that as an instant opinion, the BSC often does not carry too much content, and users often express their opinions in a short and concise way.
Besides, the statistical results of the number of 'likes' show that users have strong opinions to interact with other users when browsing, and the proportion of highly recognized BSCs is not low, which proves the value of BSC text laterally.

\subsubsection{Bullet screen comment and plot}
\label{sec:bsc_plot}
As real-time comments are sent by viewers when they watch videos, the number of BSCs can usually reflect the user's attention to specific sections in the video. 
Furthermore, these segments are often the key plots in TV dramas.
Therefore, we try to analyze the relationship between the popularity of BSCs in the video segment and the plot clip annotation information in \TVD.
We first use the sliding window to count the number of BSCs in intervals of 10 seconds, and then use the peak-detecting algorithm~\citep{peakdet2006bio} to obtain the top 5 hot segments (i.e., key plots) in each episode of TV dramas after smoothing the data, as shown in Fig.~\ref{fig:bsc_example}.
After obtaining these key plot clips, we use the plot clip annotation information in \TVD for matching.
When the overlapping rate of the annotated plot interval and the hotspot interval exceeds 80\%, it can be considered as hitting the key plot, the coverage rate of the key plot in \TVD is 83.3\%; when the overlap rate is set to 50\%, the key plot in \TVD The coverage rate is 97.7\%.

\begin{table*}
\centering
\small
\begin{tabular}{lccccccc}
\toprule
 & MLE & Miller\_Madow & Jackknife & Horvitz\_Thompson & Chao\_Shen & Wolpert\_Wolf & NSB \\
\midrule
plot & 7.05 & 7.07 & 7.07 & 7.12 & 7.09 & 7.18 & 7.09 \\
BSC & 6.64 & 6.65 & 6.65 & 6.71 & 6.68 & 6.73 & 6.66 \\
\bottomrule
\end{tabular}
\caption{\label{tab:entropy}
The entropy of plot and BSC text in \TVD. 
The units of values in the table are nat, 1 nat = $1/\mathrm{ln}2 \approx$ 1.44 bits.
}
\end{table*}

As the text was obtained by manual editing, the plot text and the BSC text are very different in expression habits, forms, and attention perspectives.
This is mainly due to the fact that plot annotators have stronger professionalism and focus on necessary narrative plot information.
However, the BSC comes from the viewers. 
Although the language usage of BSC is random, the pattern is uniform, and the focus points are also relatively consistent, focusing on actors and specific plots.
In order to analyze and then measure this difference, we wish to fully study the difference in the distribution of languages in these two texts (i.e., plot text and BSC text).
Specifically, referring to the work of~\citep{entropy2022acl}, we used various entropy estimators to calculate the plot and BSC text respectively (20,000 pieces each), and the results are shown in Tab.~\ref{tab:entropy}.

Specifically, each text entropy is calculated as follows:
\begin{itemize}
    \item Maximum Likelihood Estimation (MLE): This is the usual naive way to calculate the entropy of a sample--just measure the observed probabilities over outcomes and plug those in as if they were the true values (i.e. we have a model of the distribution that maximises the likelihood of our observations).
    \item Miller-Madow: Applies a first-order Taylor correction to MLE.
    \item Jackknife: Using subsamples of our sample, it attempts to project the entropy estimate for the whole sample.
    \item Horvitz-Thompson: Takes into account the likelihood of an outcome occurring at all in our sample--less-frequent outcomes contribute more to the entropy in order to compensate for MLE's negative bias.
    \item Chao-Shen: Modifies Horvitz-Thompson to use a coverage-reduced estimate of the probability of the outcomes. Empirically better than it.
    \item Wolpert-Wolf: First Bayesian entropy estimator with a Dirichlet prior.
    \item Nemenman-Shafee-Bialek (NSB): Improves upon Wolpert-Wolf.
\end{itemize}

From the results shown in Tab.~\ref{tab:entropy}, we can observe that the entropy of BSC text is consistently much lower than that of episodic text across all estimators.
In particular, when we select much more BSC texts (2,000,000 pieces) to measure entropy, the obtained results are basically consistent with the existing results (20,000 pieces).
We believe that this intuitively indicates that the internal language usage and expression of the BSC text tend to be more unified than the plot text, reflecting that users often have more fixed words or formats when posting BSCs.
Compared with BSCs, plot texts need to pay more attention to narrative quality, and the ways of words and sentences will be more diverse.

\section{Experiment Setup}
\label{sec:app_experiment_setup}

We will describe the dataset splitting, parameters, and two-stage (pre-training and fine-tuning) process of the framework in this section.

\subsection{Dataset splitting}
\label{sec:app_dataset_splitting}

We split training, validation, and test data in 8:1:1 for all downstream tasks (i.e., genre classification, cross-modal retrieval, and plot text ).
In addition, to ensure that entities, such as characters, appearing in the validation and test sets are already present in the training set, we make the plot clips of each TV drama appear uniformly but randomly in the training, validation, and test sets.

\subsection{Pre-training settings}
\label{sec:app_pretrain_setting}

In the three experiments, we used vision and text encoders are vit\_base\_p16\_224\footnote{\url{https://huggingface.co/google/vit-base-patch16-224}} and bert-base-chinese\footnote{\url{https://huggingface.co/bert-base-chinese}} respectively.
In the image-text pre-training model, we use the AdamW optimizer~\citep{adamw2017axriv} with the learning rate set to 1e-5 for the pre-training model.
The warm-up ratio is set to 10\%, and the learning rate decays linearly to 0 after 10\% of the total training steps.
The batch size, hidden size, and number of heads are set to 4096, 768, and 12, respectively.
We pre-trained our model for 100K steps on 8 NVIDIA A100 GPUs, which takes around 5 days.
In the video-text pre-training model, since the video representation is obtained by inputting four frames of the video into the image encoder in our framework, we have completely followed the settings of all image-text pre-training models. 
This model is also pre-trained for 100K steps on 8 NVIDIA A100 GPUs, which takes about 8 days.

\subsection{Fine-tuning settings}
\label{sec:app_finetune_setting}

For the downstream tasks, we perform grid searches over the learning rates and image resolutions. 
Finally, the learning rate of the three models is uniformly set to 1e-5, the image resolution is 384, and the max text length is 50.

\section{More Empirical Results}
\label{sec:app_additional_exp_res}

\subsection{Unimodal performance in genre classification}
\label{sec:app_unimodal_genre_cls}

\begin{table*}
\centering
\begin{tabular}{llcccccc}
\toprule
\multirow{2}{*}{\textbf{Modality}} & \multirow{2}{*}{\textbf{Source}} & \multicolumn{3}{c}{\textbf{Micro}} & \multicolumn{3}{c}{\textbf{Macro}} \\
&  &  P & R & F1 & P & R & F1 \\
\midrule
\multirow{2}{*}{Text} & BSC & 80.2 & 70.3 & 75.0 & 77.4 & 59.1 & 65.2 \\
 & Plot & 88.1 & 88.5 & 88.3 & 78.9 & 79.4 & 79.1 \\
\midrule
Image & Frame & 43.8 & 51.6 & 47.7 & 39.4 & 42.5 & 40.9 \\
\midrule
Video & Clip & 49.3 & 58.6 & 58.7 & 41.3 & 44.6 & 42.9 \\
\bottomrule
\end{tabular}
\caption{\label{tab:app_genre_cls}
The results for unimodal data in genre classification of \TVD.
}
\end{table*}

Tab.~\ref{tab:app_genre_cls} lists the results of genre classification based on TV drama data of different modalities.
Observing the performance gap between data with different modalities reveals that the modality of the data has a great impact on genre classification when processing the same plot clip.
Specifically, in the genre classification task, the performance of visual signals is significantly lower than that of text signals such as plots or BSCs.
On the one hand, this is because the content of the text signal has been manually edited, which can summarize the plot characteristics with high quality, and can more directly reflect the genre of TV dramas compared with the visual signal.
On the other hand, existing text encoders generally outperform visual encoders in semantic representation capabilities.

There are also differences in the performance of text signals from the three different sources.
Among them, TV drama abstract has the worst support for genre prediction, mainly due to the small number of tag samples in some genres.
This directly makes it difficult for the model to fully capture the genre characteristics in the case of few shots, making some categories perform poorly, resulting in the Macro results being significantly lower than the Micro results.
Genre classification based on BSC and plot text shows high performance, indicating the data value of the above text in plot analysis.
In addition, the genre classification performance based on BSC is close to that of manually edited plots in Precision and far lower in Recall, indicating that BSC content can guide the understanding of genres to a certain extent, but some noise samples lead to the inability to recall all samples.

\subsection{The number of frames in the video}
\label{sec:app_number_frame}

\begin{table*}[t]
\centering
\begin{tabular}{c|cccccc}
\toprule
\textbf{\#frames} & \textbf{VR@1} & \textbf{VR@5} & \textbf{VR@10} & \textbf{TR@1} & \textbf{TR@5} & \textbf{TR@10} \\
\midrule
 2 & 1.8 & 7.8 & 14.6 & 1.8 & 7.8 & 15.8 \\
 4 & 1.8 & 8.9 & 15.5 & 2.3 & 9.4 & 16.9 \\
 6 & 2.0 & 10.5 & 18.0 & 2.4 & 12.2 & 19.4 \\
\bottomrule
\end{tabular}
\caption{\label{tab:frames}
The video-text retrieval result of \TVD with different numbers of frames.
}
\end{table*}

In this paper, we achieve video feature acquisition by uniformly intercepting frames in the video and stitching them in the ViT.
In order to consider the quality of the video feature and its performance in downstream tasks with the different numbers of intercepted frames, we set different intercepted numbers and compared them in cross-modal retrieval. 
The results are shown in Tab.~\ref{tab:frames}.
Through observation, it can be found that with the increase in the number of intercepted frames in the video, the performance of the model has been greatly improved in terms of VR and TR indicators.
This shows that when processing video, considering as much video signal information as possible can effectively improve its performance in downstream tasks.

\subsection{Bullet screen comment generation}
\label{sec:app_gt}

\begin{table*}
\centering
\scriptsize
\begin{tabular}{clccccc}
\toprule
\textbf{\#BSC for reference} & \textbf{Settings} & \textbf{BLEU@1} & \textbf{BLEU@2} & \textbf{METEOR} & \textbf{CIDEr} \\
\midrule
\multirow{2}{*}{BSC-1} & Complete model & 9.1 & 3.9 & 4.3 & 7.6 \\
& \textit{-w/o BSC pre-train}  & 10.1 & 4.3 & 4.4 & 8.5 \\
\midrule
\multirow{2}{*}{BSC-3} & Complete model & 25.5 & 11.8 & 7.2 & 5.9 \\
& \textit{-w/o BSC pre-train}  & 24.7 & 11.0 & 7.2 & 6.5 \\
\midrule
\multirow{2}{*}{BSC-5} & Complete model & 36.6 & 17.8 & 8.8 & 5.4 \\
& \textit{-w/o BSC pre-train}  & 32.6 & 15.6 & 8.7 & 5.7 \\
\bottomrule
\end{tabular}
\caption{\label{tab:BSC_generation}
The results of BSC generation.
}
\end{table*}

\begin{table*}
\centering
\tiny
\begin{tabular}{cll}
\toprule
\textbf{Video} & \multicolumn{2}{c}{\textbf{Text}} \\
\midrule
\multirow{4}{*}{\includegraphics[width=0.3\linewidth]{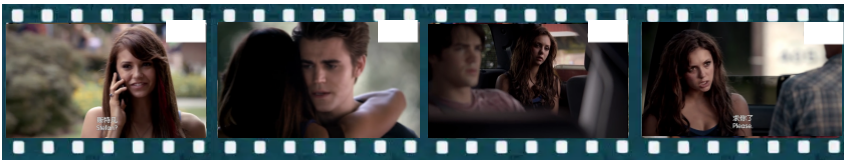}} & \multirow{2}{*}{GT:} & \begin{CJK*}{UTF8}{gbsn}塞勒斯到艾琳娜的学校找到了艾琳娜，他以斯特凡的身份和艾琳娜谈论了达蒙。\end{CJK*} \\
 & & Cyrus went to Elena's school to find Elena, and he talked to Elena about Damon as Stefan. \\
 & \multirow{2}{*}{GR:} & \begin{CJK*}{UTF8}{gbsn}斯特凡到艾琳娜那里去找艾琳娜，艾琳娜告诉他自己不会有任何事情。\end{CJK*} \\
 & & Stefan went to find Elena, and Elena told him that nothing would happen to her. \\
 & & \\
\multirow{4}{*}{\includegraphics[width=0.3\linewidth]{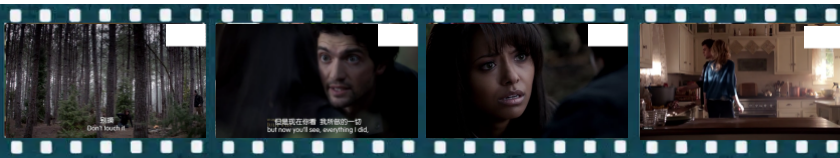}} & \multirow{2}{*}{GT:} & \begin{CJK*}{UTF8}{gbsn}谢恩找到了邦妮，他要邦妮小心复活的塞拉斯。\end{CJK*}  \\
 & & Shane finds Bonnie and tells her to watch out for the resurrected Silas. \\
 & \multirow{2}{*}{GR:} & \begin{CJK*}{UTF8}{gbsn}艾琳娜和达蒙见到了马特，并要他们多注意。\end{CJK*} \\
 & & Elena and Damon meet Matt and ask them to pay more attention. \\
 & & \\
\multirow{4}{*}{\includegraphics[width=0.3\linewidth]{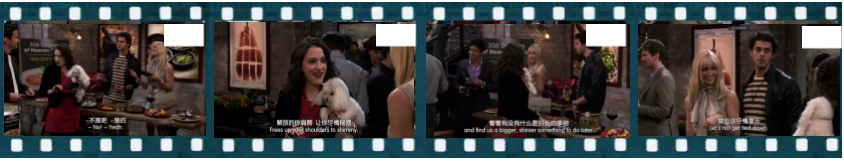}} & \multirow{2}{*}{GT:} & \begin{CJK*}{UTF8}{gbsn}在那里麦克斯和凯若琳遇到了她们英俊的邻居。\end{CJK*}  \\
 & & There Max and Caroline meet their handsome neighbor. \\
 & \multirow{2}{*}{GR:} & \begin{CJK*}{UTF8}{gbsn}麦克斯和凯若琳在餐厅见面，她们正在准备约会。\end{CJK*} \\
 & & Max and Caroline meet at a restaurant. They're getting ready for a date. \\
 & & \\
\multirow{4}{*}{\includegraphics[width=0.3\linewidth]{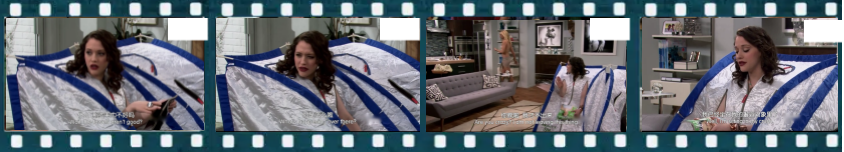}} & \multirow{2}{*}{GT:} & \begin{CJK*}{UTF8}{gbsn}这点让凯若琳很失望，她准备去找吃的。\end{CJK*}  \\
 & & This disappointed Caroline, she was going to find something to eat. \\
 & \multirow{2}{*}{GR:} & \begin{CJK*}{UTF8}{gbsn}凯若琳表示自己很不高兴，她要麦克斯一起出去。\end{CJK*} \\
 & & Caroline says she's upset and wants Max to go out with her. \\
\bottomrule
\end{tabular}
\caption{\label{tab:add_plot_generation_case}
More cases of plot text generation in \TVD. GT is short for ground truth. GR is short for generation results.
}
\end{table*}

In order to verify the performance of BSC pre-training in BSC generation and examine the results of introducing BSC at the same time point as evaluation references, we conducted relevant experiments, and the results are shown in Tab.~\ref{tab:BSC_generation}.
It can be found that when more BSC at the same time point are gradually introduced, the performance of the model is gradually improved, and BSC pre-training also shows a gain effect.

\subsection{More generation results}
\label{sec:app_generation_res}

\begin{figure}[ht]
\centering
\includegraphics[width=0.6\textwidth]{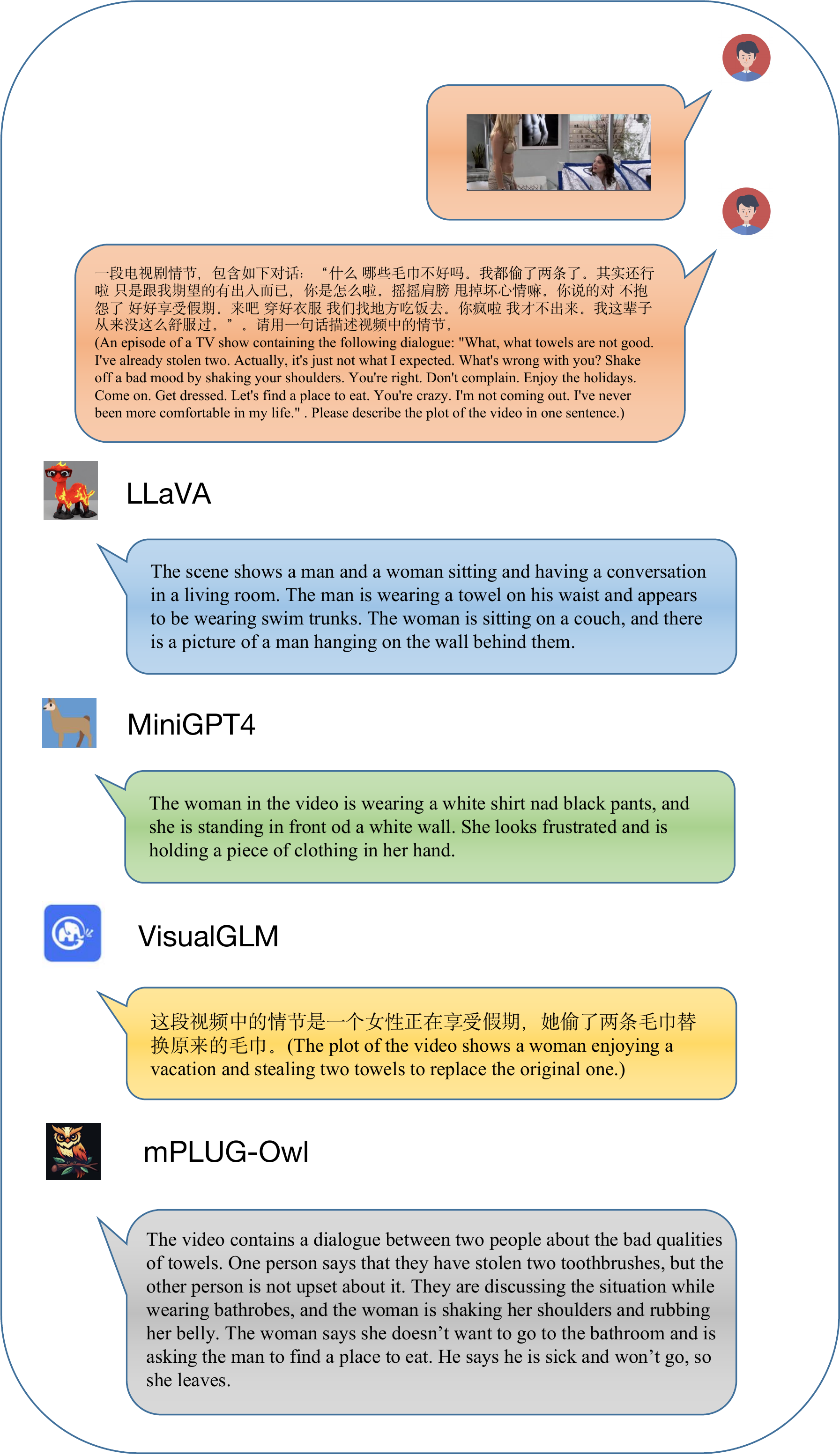}
\caption{
An example of plot text generation results for multimodal LLMs.
}
\label{fig:mmllm}
\end{figure}

We list more plot text generation results here, as shown in Tab.~\ref{tab:add_plot_generation_case}.
Through observation, it can be clearly found that the video-text pre-training model, which can perform well on the public website dataset~\citep{pretrainvl12020arxiv}, does not achieve ideal results in plot generation.
This is mainly due to the difference between the TV drama data and the publicly collected dataset, i.e., the TV drama data contain more explicit entity names (first two rows in Tab.~\ref{tab:add_plot_generation_case}) and long-distance temporal relationships (last two rows in Tab.\ref{tab:add_plot_generation_case}), and the existing model is difficult to deal with such information.
Certainly, we can also find that the video-text pre-training model we used can generate more semantic information with situation context and mental state based on visual signals, indicating that the existing multimodal understanding model can try to understand the information hidden behind visual signals.

Recently, a large number of multimodal large language models (LLMs) (e.g., LLaVA\footnote{\url{https://llava.hliu.cc/}}, MiniGPT-4\footnote{\url{https://huggingface.co/spaces/Vision-CAIR/minigpt4}}, visualGLM\footnote{\url{https://huggingface.co/spaces/lykeven/visualglm-6b}}, and mPLUG-Owl\footnote{\url{https://modelscope.cn/studios/damo/mPLUG-Owl/summary}}) have achieved outstanding performance on multimodal data understanding tasks (especially images, and videos).
We selected several well-known open-source multimodal LLMs to perform the task of plot text generation in zero-shot setting. The experimental results are shown in Fig.~\ref{fig:mmllm}.
The generation results are not satisfactory. Most of the models pay more attention to the captions of the input visual signal. This may be due to the problem of understanding the task instructions, and the visual signal processing and thought chain of the existing multi-modal LLMs inferring ability is not strong enough.
Therefore, the plot comprehension and generation tasks are still challenging tasks.


\section{Limitations}
\label{sec:limitations}
\paragraph{Dataset.}
Due to the high cost of collecting and annotating data, there are currently only 83 TV dramas in \TVD.
As analysed in original paper, the genre distribution of \TVD is unbalanced.
The text data of \TVD (i.e., the plot text and Bullet Screen Comments) is in Chinese only, and heavily reflect Chinese culture and expression habits. Therefore, \TVD may exhibit gaps between real-world data from other social backgrounds.
Also, although we adhere to strict standards in our data collection pipeline, we are not able to manually check all the samples, so \TVD may contain a small percentage of annotation errors.

\paragraph{Framework.}
To reduce the environmental impact and to highlight scalability, the framework implemented in this study adopts the very popular techniques rather than the most advanced ones.
As identified in the original paper, our framework may aggregate existing distribution bias in the dataset, which should be treated very carefully.
In the task of Plot Text Generation, we did not conduct manual evaluation as our budget has run out during the creation of \TVD (annotation expense, copyright overhead, etc.).
However, we should remind readers that human judgement is still needed in some cases.

\paragraph{Experiments.}
Our framework, together with \TVD, could be used to explore a wide range of tasks, but due to time limit we only tested three most representative ones.
We made several counter-intuition observations. Although we have provided reasonable guesses on their reasons, we admit a very in-depth investigation still requires a considerable amount of work.

In general, despite the above limitations, we believe that our dataset, our framework, and our empirical findings make an important addition to the multi-modal learning, or more generally, the machine intelligence research field. 

\section{Ethics Statement}
\label{sec:ethics}

\paragraph{Data life-cycle.}
Our data annotation and release plan has been examined and approved by relevant institutional committees.
All annotators participating the creation of \TVD have agreed relevant terms and received a set of pre-annotation training.
They were provided with remuneration that is significantly higher than the local minimum wage.
At different stages of data collection, we complied with corresponding legal and ethical requirements.
In order to protect the privacy of TV drama viewer, we have also anonymised all data (especially Bullet Screen Comments), ensuring no information that can uniquely identify users is excluded from our dataset and research process.
We have obtained all data copyrights relevant to this paper.
We have prepared detailed instructions, licenses, and data use agreement for future users of \TVD, aiming to minimise the possibility of any kind of misuse (e.g., to train a malicious Bullet Screen Comment writer).

\paragraph{Data bias.}
Most of the TV dramas included in \TVD are produced in a fictional context, and the relevant content may have the specific bias introduced by the original creators.
Meanwhile, the filming techniques and editing strategies commonly used in film and television photography will also highlight some dramatic plots and may have a preference for some content.
In \TVD,  the Bullet Screen Comments carry the opinions of the user, which may therefore contain personal bias.
The above bias may be inherited or even aggregated by the downstream algorithms.
Thus, we urge future users of \TVD to pay special  attention to this matter.

\end{document}